\documentclass{article}

\usepackage{makecell}
\usepackage{appendix}
\usepackage[preprint]{neurips_2021}

\usepackage[utf8]{inputenc} 
\usepackage[T1]{fontenc}    
\usepackage{hyperref}       
\usepackage{url}            
\usepackage{booktabs}       
\usepackage{amsfonts}       
\usepackage{nicefrac}       
\usepackage{microtype}      
\usepackage{xcolor}         

\usepackage{graphicx}
\usepackage{subfigure}
\usepackage{amssymb}
\usepackage{mathtools}

\usepackage{algorithm}
\usepackage{algorithmic}

\usepackage{amsmath}
\usepackage{amsthm}

\newtheorem{theorem}{Theorem}

\newtheorem{lemma}{Lemma}

\newtheorem{remark}{Remark}
\newtheorem{assumption}{Assumption}
\newtheorem{corollary}{Corollary}

\title{A New Adaptive Gradient Method with Gradient Decomposition}

\author{%
  Zhou Shao \\
  Center for Data Science\\
  Peking University\\
  Beijing, China \\
  \texttt{shaozhou@pku.edu.cn} \\
   \And
   Tong Lin \\
   Key Lab. of Machine Perception (MOE) \\
School of EECS, Peking University\\
Beijing, China \\
Peng Cheng Laboratory\\
Shenzhen, China \\
\texttt{lintong@pku.edu.cn}
}

\begin{document}

\maketitle

\begin{abstract}

Adaptive gradient methods, especially Adam-type methods (such as Adam, AMSGrad, and AdaBound), have been proposed to speed up the training process with an element-wise scaling term on learning rates. However, they often generalize poorly compared with stochastic gradient descent (SGD) and its accelerated schemes such as SGD with momentum (SGDM). In this paper, we propose a new adaptive method called DecGD, which simultaneously achieves good generalization like SGDM and obtain rapid convergence like Adam-type methods. In particular, DecGD decomposes the current gradient into the product of two terms including a surrogate gradient and a loss based vector. Our method adjusts the learning rates adaptively according to the current loss based vector instead of the squared gradients used in Adam-type methods. The intuition for adaptive learning rates of DecGD is that a good optimizer, in general cases, needs to decrease the learning rates as the loss decreases, which is similar to the learning rates decay scheduling technique. Therefore, DecGD gets a rapid convergence in the early phases of training and controls the effective learning rates according to the loss based vectors which help lead to a better generalization. Convergence analysis is discussed in both convex and non-convex situations. Finally, empirical results on widely-used tasks and models demonstrate that DecGD shows better generalization performance than SGDM and rapid convergence like Adam-type methods.

\end{abstract}

\section{Introduction}

Consider the following stochastic optimization problem:
\begin{equation}
	\label{s}
	\min_{x\in\mathcal{X}}f(x)\coloneqq\mathbb{E}_{S\sim P}f(x;S)=\int_{S}f(x;s)\mathrm{d}P(s),
\end{equation}
where $S\sim P$ is a random variable, $f(x;s)$ is the instantaneous loss parameterized by $x$ on a sample $s\in S$, $\mathcal{X}\subseteq\mathbb{R}^d$ is a closed convex set. This problem is a common learning task in machine learning and deep learning. Many efforts have been spent on proposing stochastic optimization methods to solve this problem. Stochastic gradient descent (SGD) is the dominant first-order method for the above problem \cite{8,9,4}. SGD is often trained in the form of mini-batch SGD in order to meet the requirements of computing power, and achieve better generalization performance \cite{21,22}. However, SGD has the following main drawbacks. First, SGD chooses the negative gradients of loss functions as descent directions which would yield a slow convergence near the local minima. Second, SGD scales the gradients uniformly in all directions which may yield poor performance as well as limited training speed. Last but not least, when applied to machine learning and deep learning tasks, SGD is painstakingly hard to tune the learning rates decay scheduling manually. However, one has to decay learning rates as the algorithm proceeds in order to control the variances of stochastic gradients for achieving convergence due to the high-dimensional non-convexity of machine learning and deep learning optimization problems.

To tackle aforementioned issues, considerable efforts have been spent and several remarkable variants have been proposed recently. Accelerated schemes and adaptive methods are two categories of dominant variants. Accelerated schemes, such as Nesterov’s accelerated gradient (NAG) \cite{12} and SGD with momentum (SGDM) \cite{11}, employ momentum to adjust descent directions which can help achieve faster convergence and better generalization performance than other variants. However, they also suffer from the third drawback of SGD so that one need to spend many efforts on tuning and decaying learning rates manually. On the other hand, adaptive methods aim to alleviate this issue which automatically decay the learning rates and scale them nonuniformly. The first prominent algorithm in this line of research is AdaGrad \cite{13}, which divides element-wisely accumulative squared historical gradients. AdaGrad performs well when gradients are sparse, but its performance degrades in dense or non-convex settings which is attributed to the rapid decay in learning rates. Towards this end, several methods proposed scale gradients down by square roots of exponential moving averages of squared historical gradients (called EMA mechanism) which focus on only the recent gradients. This mechanism is very popular and some famous variants, including AdaDelta \cite{14}, RMSprop \cite{15} and Adam \cite{16}, are based on it. Particularly, Adam is a combination of momentum and EMA mechanism which converges fast in the early training phases and is easier to tuning than SGD, becoming the default algorithms leveraged across various deep learning frameworks.

Despite Adam's popularity, there also have been concerns about their convergence and generalization properties. In particular, EMA based methods may not converge to the optimal solution even in simple convex settings \cite{3} which relies on the fact that effective learning rates of EMA based methods can potentially increase over time in a fairly quickly manner. For convergence, it is important to have the effective learning rates decrease over iterations, or at least have controlled increase \cite{Yogi}. Moreover, this problem persists even if the learning rate scheduling (decay) is applied. Recently, considerable efforts have been spent on improving EMA based methods \cite{3,17,Yogi,PAdam,RAdam,23} to narrow the generalization gap between EMA based methods with SGD. However, one pursuing the best generalization ability of models has to choose SGD as the default optimizer rather than Adam based on the fact that there is not enough evidence to show that those Adam-type methods, which claim to improve the generalization ability of the EMA mechanism, can get close to or even surpass SGD in general tasks. Therefore, a natural idea is whether it is possible to develop a new adaptive method different form EMA based methods, which can overcome aforementioned issues and obtain even better generalization than SGD. However, to the best of our knowledge, there exists few efforts on proposing new adaptive mechanisms whose starting points are different form EMA mechanism. 

\paragraph{Contributions}

In the light of this background, we list the main contributions of our paper.
\begin{itemize}
	\item[$\bullet$] We propose a new adaptive method, called DecGD, different from Adam-type methods. DecGD decomposes gradients into the product of two terms including surrogate gradients and loss based vectors. Our method achieves adaptivity in SGD according to loss based vectors with the intuition that a good optimizer, in general cases, needs to decrease the learning rates as the loss decreases, which is similar to the learning rates decay scheduling technique. DecGD overcomes aforementioned drawbacks of SGD and achieve comparable and even better generalization than SGD with momentum.
	
	\item[$\bullet$] We theoretically analyze the convergence of DecGD in both convex and non-convex settings.
	
	\item[$\bullet$] We conduct extensive empirical experiments for DecGD and compare with several representative methods. Empirical results show that DecGD is robust to hyperparameters and learning rates. Moreover, our method achieves fast convergence as Adam-type methods and shows the best generalization performance in most tasks.
\end{itemize}

\paragraph{Related Work}

The literature in stochastic methods is vast and we review a few very closely related work on improving SGD or Adam. These proposed methods can simply be summarized into two families: EMA based methods and others. For EMA based methods, many efforts have been spent on closing the generalization gap between Adam and SGD family. AMSGrad \cite{3} controls the increasing of effective learning rates over iterations, AdaBound \cite{17} clips them, Yogi \cite{Yogi} considers the mini-batch size, PAdam \cite{PAdam} modifies the square root, RAdam \cite{RAdam} rectifies the variance of learning rates, AdamW \cite{AdamW} decouples weight decay from gradient descent, AdaBelief \cite{23} centralizes the second order momentum in Adam. To our best knowledge, there exists several methods achieving different adaptivity from Adam-type methods in SGD. AdaGD \cite{27} focuses on the local geometry and use the gradients of the latest two steps to adaptively adjust learning rates, which has a convergence only depending on the local smoothness in a neighborhood of the local minima. Besides, AEGD \cite{19} is closest to our work which uses the same composite function of the loss with ours: $\sqrt{f(x)+c}$ where $f$ is the loss and $c$ is a constant s.t. $f(x)+c>0$ for all $x$ in the feasible region. However, the intuition for AEGD which is far different from our method is to realize a gradient descent with the stable energy which is defined as the above composite function of the loss. Note that the energy of AEGD equals the definition just in the first step and updates in a monotonous way as the algorithm proceeds. Hence, the stable energy seems meaningless because that it unconditionally decrease over iterations.

\paragraph{Notation}
For a vector $\theta \in \mathbb{R}^d$, we denote its $i$-th coordinate by $\theta_i$. We use $\theta_t$ to denote $\theta$ in the $t$-th iteration and use $\theta_{t,i}$ for the $i$-th coordinate of $\theta$ in the $t$-th iteration. Furthermore, we use $||\cdot||$ to denote $l_2$-norm and use $||\cdot||_{\infty}$ to denote $l_{\infty}$-norm. Given two vectors $v,w\in\mathbb{R}^d$, we use $vw$ to denote element-wise product and use $v^2$ to denote element-wise square. We use $\frac{w}{v}$ to denote element-wise division.



\section{DecGD}

As summarized before, one with aim to improve SGD performance in machine learning and deep learning tasks should consider the following three directions: improving descent directions, a non-uniform scale, and a combination with the learning rate decay scheduling technique. Momentum based methods pay attention to the first aspect, while adaptive methods make efforts to achieve adaptivity with element-wise operations motivated by the second direction. In particular, the dominant adaptive methods are Adam-type methods such as Adam, AdaBound and AdaBelief which employ the second raw moment or second central moment of stochastic gradients, also called EMA mechanism, to achieve adaptivity. However, although many efforts have been spent on the generalization ability, there is not enough evidence showing that Adam-type methods could generalize better than SGD. Therefore, SGD may need a new and different adaptivity. 

In the light of this background, we propose a new adaptive variant of SGD, called DecGD, which decomposes gradients into the product of two terms including surrogate gradients and loss based vectors. DecGD achieves new adaptivity in SGD with the loss based vectors and the pseudo-code is shown in Algorithm \ref{alog1}.

\begin{algorithm}
	\caption{DecGD (default initialization: $c=1$, $\gamma=0.9$, $\eta=0.01$, AMS=False)}
	\label{alog1}
	\begin{algorithmic}
		\STATE{\bfseries Input:} $x_1 \in \mathcal{X}\subseteq\mathbb{R}^d$, learning rate $\eta$, $c$, momentum $\gamma$, AMS
		\STATE{\bfseries Initialize} $x_{0,i}=0$, $m_{0,i}=0$, $v_{0,i}=\sqrt{f(x_1)+c}$, $i=1,2,\cdots,d$
		\FOR{$t=1$ {\bfseries to} $T$}
		\STATE $\nabla g\leftarrow \nabla f(x_t)/2\sqrt{f(x_t)+c}$\ \ \ \ \ \  \ \ \ \ \ \ \ \ \ \ \ \ \ \ (compute the scaled gradients)
		\STATE $m_{t}\leftarrow \gamma m_{t-1} + \nabla g$\ \ \ \ \ \ \ \ \ \ \ \ \ \ \ \ \ \ \ \ \ \ \ \ \ \ \ \ \ \ \ \ \ \ \  (update the first order momentum)
		\STATE $v_{t}\leftarrow v_{t-1} + m_t(x_{t}-x_{t-1})$\ \ \ \ \ \ \ \ \ \ \ \ \ \ \ \ \ \ \ \  \ (update the loss based vector)
		\STATE $v^*=\min\{v^*,v_t\}$ \textbf{if AMS else} $v^*=v_t$\ \ \  \ (choose a monotonically decreasing $v$ or not)
		\STATE $x_{t+1}\leftarrow x_{t}-2\eta v^*m_t$\ \ \ \ \ \ \ \ \ \ \ \ \ \ \ \ \ \ \ \ \ \ \ \ \ \ \ \ \ \ \ \ \ (update the parameters)
		\ENDFOR
	\end{algorithmic}
\end{algorithm}

\paragraph{Intuition}
The intuition for DecGD is that the loss can help to adjust learning rates over iterations. As the algorithm proceeds, typically one need to adjust the learning rates for convergence. Thus, the learning rate decay scheduling technique is often applied into the training process. Note that both momentum based methods and adaptive methods benefit from the combination with the learning rate decay scheduling technique. Hence, adjusting learning rates according to the loss information is feasible. DecGD employs a decomposition of gradients to access the loss information.

\paragraph{Gradient Decomposition}
Consider a composite function of the loss $f(x)$:
\begin{equation}
	\label{composite}
	\begin{aligned}
		&g(x)=\sqrt{f(x)+c},
	\end{aligned}
\end{equation}
where the objective loss $f(x)$ is a lower bounded function and $c>0$ is a constant s.t. $f(x)+c>0$, $\forall x\in\mathcal{X}\subseteq\mathbb{R}^d$. Take the derivative of $g(x)$ and we can decompose the gradient $\nabla f(x)$ into the product of two terms
\begin{equation}
	\label{decomposite}
	\begin{aligned}
		& \nabla f(x)=2g(x)\nabla g(x),
	\end{aligned}
\end{equation}
where $\nabla f(x)$ and $\nabla g(x)$ are the gradient of $f(x)$ and $g(x)$ respectively. Note that $g(x)$ which has the same monotonicity with $f(x)$ includes the current information of the loss, and $\nabla g(x)$ is a scaled version of $\nabla f(x)$ with a factor $2g(x)$ which is a constant for a certain $x$. Thus, $-\nabla g(x)$ is also a descent direction because we have $-\nabla g(x)\nabla f(x)=-(\nabla f(x))^2/2g(x)<0$ where $g(x)>0$, $\forall x\in\mathcal{X}$. In conclusion, we decompose the gradient of the loss $f(x)$ into a surrogate gradient $\nabla g(x)$ which is a descent direction for optimizing $f(x)$ and a loss scalar $g(x)$. 

\paragraph{Update Rules}
Based on the above decomposition, we show the update rule of DecGD. To determine an optimizer, all we need is calculating the learning rate (step size) and searching the update direction. For example, the update scheme of vanilla SGD is: 
\begin{equation}
	\label{sgd}
	\begin{aligned}
		& x_{t+1}=x_t - \eta\nabla f(x_t),
	\end{aligned}
\end{equation}
where $x_{t+1}$, $x_t\in\mathbb{R}^d$, $\eta$ is a constant learning rate, $\nabla f(x_t)$ is the gradient at the time step $t$ and $-\nabla f(x_t)$ is the steepest descent direction. 

First, we consider the update direction of DecGD. As mentioned above, $-\nabla g(x)$ is the descent direction which is actually the scaled vector of the steepest descent direction. We employ the momentum to achieve acceleration: 
\begin{equation}
	\label{m}
	\begin{aligned}
		&m_{t}= \gamma m_{t-1} + \nabla g,
	\end{aligned}
\end{equation}
where $m_{t+1},m_t\in\mathbb{R}^d$, a constant $\gamma\in(0,1)$ control the exponential decay rate and the initialization $m_{0,i}=0$, $i=1,2,\cdots,d$. The above formula shows how to update the direction of DecGD. 

We next consider calculating the learning rate of DecGD. In machine learning and deep learning optimization problems, the learning rate at a certain time step is often a constant rather than calculated by traditional line search for the sake of computing cost. To avoid the second issue of SGD, the learning rate often multiplies a vector element-wisely for adaptivity. DecGD, motivated by the learning rate decay scheduling technique, employs a loss based vector which comes from the loss scalar in the above gradient decomposition and has the following update rule
\begin{equation}
	\label{v}
	\begin{aligned}
		v_{t}=v_{t-1} &+ m_t(x_t-x_{t-1}),\\
		v^*=\min\{v^*,&v_t\}\ \textbf{if AMS else}\ v^*=v_t,
	\end{aligned}
\end{equation}
where the initial vector $v_{0,i}=g(x_0)$, $i=1,2,\cdots,d$. Note that applying the loss scalar $g(x)$, which is the second term of the gradient decomposition, directly results in two disadvantages: First, $g(x)$ is a scalar for a certain $x$ so that we fail to achieve a non-uniform adaptive learning rate. Besides, if so, DecGD would exactly equal to SGD or SGD with momentum. Therefore, we employ the first order Taylor polynomial to approximate $g(x)$ linearly to achieve the approximate loss. In detail, we start with a vector with the initial value $g(x_0)$ and update it element-wisely to get a non-uniform loss based vector $v$ in various directions. The update rule \ref{v} can be viewed as the momentum version of the first order Taylor polynomial. With the same aim as AMSGrad, DecGD provides a switch on whether to rectifies $v_t$ to ensure that $v$ would not increase in a fairly vast way. 

DecGD employs the following scheme to update parameters based on the above decomposition
\begin{equation}
	\label{decgd}
	\begin{aligned}
		& x_{t+1}=x_t - 2\eta v^* m_t.
	\end{aligned}
\end{equation}
Finally, we note that the computing complexity and the time complexity of DecGD are both $O(d)$ which is same with Adam. However, DecGD has one less parameter than Adam (the 'AMS' in Adam is whether to enable AMSGrad). 

\paragraph{Relevance and Difference}
With the same aim to achieve adaptivity in SGD, DecGD and Adam-type methods all employ element-wise operations to scale the learning rates non-uniformly and apply momentum to update directions. Differently, Adam-type methods have been working on using the square gradients to approximate the Hessian to obtain second-order information. In particular, AdaBelief \cite{23}, which centralizes second order momentum, considers the variance in gradients. In fact, it is a kind of approximation of Hessian to achieve second order information. AdaGD \cite{27} considers the local Lipschitz constant which is similar to AdaBelief actually. However, it is very different that DecGD considers a zero order information motivated by the practical application. It seems that we can integrate DecGD with Adam-type methods to achieve both zero order and second order approximate information for better convergence. This topic remains open.



\section{Convergence Analysis}
We discuss the convergence of DecGD in both convex and non-convex situations. The convergence under the condition of convex objective functions is showed in the online convex optimization framework \cite{13, 3, 28, 29} which is similar to Adam \cite{16}, AMSGrad \cite{3}, AdaBound \cite{17} and AdaBelief \cite{23}. Furthermore, we analyze the convergence in the stochastic non-convex optimization problem, which is similar to the previous work \cite{23,adamtype}. This situation is more in line with actual scenarios of machine learning and deep learning tasks.

\subsection{Online Convex Optimization}
In online optimization, we have a loss function $f_t:\mathcal{X}\to\mathbb{R}$. After a decision $x_t\in\mathcal{X}$ is picked by the algorithm, we have the following regret to minimize:
\begin{equation}
	R(T)=\sum_{i=0}^Tf_t(x_t)-\min_{x\in\mathcal{X}}\sum_{i=0}^Tf_t(x).
\end{equation}
The standard assumptions \cite{13, 3, 28, 29} in the setting of online convex optimization are as follows:
\begin{assumption}
	\label{a1}
	(1) $\mathcal{X}\subseteq\mathbb{R}^d$ is a compact convex set;
	(2) $f_t$ is a convex lower semi-continuous (lsc) function, $g_t\in\partial f_t(x_t)$;
	(3) $D=\max_{x,y\in\mathcal{X}}||x-y||$, $G=\max_{t}||g_t||$.
\end{assumption}
We propose the following lemma:
\begin{lemma}
	\label{l1}
	$f_t$ is a lower bounded function and $c>0$ is a constant s.t. $f_t(x)+c>0, x\in\mathcal{X}$. Let $l_t(x)=\sqrt{f_t(x)+c}$. If $f_t$ has bounded gradients, then $l_t$ has bounded gradients too and is bounded in the feasible regions.
\end{lemma}
\begin{remark}
	The above lemma shows that two terms from the decomposition of the gradient $\nabla f_t(x)$ are both bounded. In particular, the assumption (3) in the standard assumptions \ref{a1} yields $||\nabla l_t(x_t)||\le G$, $|l_t(x_t)|\le L$, $x_t\in\mathcal{X}$.
\end{remark}
Therefore, we can get the following assumptions for DecGD which are entirely yielded from the standard assumptions \ref{a1}:




\begin{assumption}
	\label{a2}
	(1) $\mathcal{X}\subseteq\mathbb{R}^d$ is a compact convex set;
	(2) $f_t$ is a convex lsc function, $l_t=\sqrt{f_t+c}$, $c>0$;
	(3) $||x-y||\le D$, $||\nabla l_t(x_t)||\le G$, $|l_t(x_t)|\le L$, $x_t\in\mathcal{X}$.
\end{assumption}
The key results are as follows:
\begin{theorem}
	\label{t1}
	Under the Assumption \ref{a2}, let $\gamma\in(0,1)$ and $\eta_t=\frac{\eta}{\sqrt{t}}$, $\eta>0$, DecGD has the following bound on the regret:
	\begin{equation}
		\begin{aligned}
			R(T)\le & \frac{LD^2\sqrt{T}}{4\eta}\sum_{i=1}^dv_{T,i}^{-1}+LD^2\sum_{t=1}^T\frac{\gamma^2}{\eta_t}+\frac{\eta G^2L(1+L)}{1-\gamma}(2\sqrt{T}-1).
		\end{aligned}
	\end{equation}
\end{theorem}
The following result falls as an immediate corollary of the above results:
\begin{corollary}
	\label{c1}
	Suppose $\gamma$ has a decay with factor $\lambda^{t-1}$ in Theorem \ref{t1}, we have
	\begin{equation}
		R(T)\le\frac{LD^2\sqrt{T}}{4\eta}\sum_{i=1}^dv_{T,i}^{-1}+\frac{LD^2\gamma^2}{\eta(1-\lambda^2)^2}+\frac{\eta G^2L(1+L)}{1-\gamma}(2\sqrt{T}-1).
	\end{equation}
\end{corollary}
\begin{corollary}
	Under the same assumptions of Theorem \ref{t1}, DecGD has the following average regret of convergence:
	\begin{equation}
		\frac{R(T)}{T}=O\Big(\sqrt{\frac{1}{T}}\Big).
	\end{equation}
\end{corollary}
\begin{remark}
	Theorem \ref{t1} implies the regret of DecGD is upper bounded by $O(\sqrt{T})$, similar to Adam \cite{16}, AMSGrad \cite{3}, AdaBound \cite{17} and AdaBelief \cite{23}. Besides, the condition in Corollary \ref{c1} can be relaxed to $\gamma_t=\gamma/\sqrt{t}$ and still ensures a regret bound of $O(\sqrt{T})$.
\end{remark}

\subsection{Stochastic Non-convex Optimization}
We discuss the convergence in the stochastic non-convex learning which is more in line with actual scenarios of machine learning and deep learning tasks than the online convex optimization. The standard assumptions \cite{23,adamtype} are as follows:
\begin{assumption}
	\label{a3}
	(1) $f$ is lower bounded and differentiable, $||\nabla f(x)-\nabla f(y)||\le L||x-y||, \forall x, y$;
	(2) The noisy gradient is unbiased, and has independent noise, i.e. $g(t)=\nabla f(\theta_t)+\zeta_t$, $\mathbb{E}\zeta_t=0$, $\zeta_t\perp\zeta_j$, $\forall t,j\in\mathbb{N}$, $t\neq j$;
	(3) At step t, the algorithm can access a bounded noisy gradient, and the true gradient is also bounded, i.e. $||\nabla f(\theta_t)||\le H$, $||g_t||\le H$, $\forall t>1$.
\end{assumption}
Similarly, the above assumptions yield the following assumptions for DecGD according to Lemma \ref{l1}:
\begin{assumption}
	\label{a4}
	(1) $f$ is lower bounded and differentiable, $||\nabla f(x)-\nabla f(y)||\le L||x-y||, \forall x, y$;
	(2) The noisy gradient is unbiased, and has independent noise, i.e. $g_t=\nabla f(\theta_t)+\zeta_t$, $\mathbb{E}\zeta_t=0$, $\zeta_t\perp\zeta_j$, $\forall t,j\in\mathbb{N}$, $t\neq j$;
	(3) At step t, the algorithm can access a bounded noisy gradient, and the true gradient is also bounded, i.e. $||l_t||\le L$, $||\hat{g}_t||\le G$, $\hat{g}_t$ is the noisy gradient of $l_t=\sqrt{f_t+c}$, $\forall t>1$.
\end{assumption}
The key result is as follows:
\begin{theorem}
	\label{t2}
	Under the Assumption \ref{a4}, let $\gamma_t<\gamma\le1$, $||v_T||_1\ge c$ and $\eta_t=\frac{\eta}{\sqrt{t}}$, $\eta>0$, DecGD satisfies
	\begin{equation}
		\begin{aligned}
			\min_{t\in[T]}\mathbb{E}\Big(||\nabla f(\theta_t)||^2\Big)\le\frac{L^2}{c\eta\sqrt{T}}\Big(C_1L^2\eta^2G^2(1+\log T)+C_2d\eta+C_3dL^2\eta^2+C_4\Big),
		\end{aligned}
	\end{equation}
	where $C_1$, $C_2$, $C_3$ are constants independent of $d$ and $T$, $C_4$ is a constant independent of $T$.
\end{theorem}
\begin{remark}
	Theorem \ref{t2} implies that DecGD has a $O(\log T/\sqrt{T})$ convergence rate in the stochastic non-convex situation which is similar to Adam-type methods \cite{adamtype,23}. Besides, Theorem 3.1 in \cite{adamtype} needs to specify the bound of each update, but DecGD needs not. The proof follows the general framework in \cite{adamtype}, and it's possible the above bound is loose. A sharper convergence analysis remains open.
\end{remark}

\section{Experiments}

In this section, we study the generalization performance of our methods and several representative optimization methods. Except SGD with momentum (SGDM), we additionally test two families of optimizers including Adam-type methods and other adaptive methods. The former includes Adam, AMSGrad, AdaBound, AdaBelief and the latter includes AEGD and our method DecGD. We conduct experiments in popular deep learning tasks for testing the performance in thestochastic situation. Particularly, several neural network structures will be chosen including multilayer perceptron, deep convolution neural network and deep recurrent neural network. Concretely, we focus on the following experiments: multilayer perceptron (MLP) on MNIST dataset \cite{4}; ResNet-34 \cite{6} and DenseNet-121 \cite{7} on CIFAR-10 dataset \cite{5}; ResNet-34 \cite{6} and DenseNet-121 \cite{7} on CIFAR-100 dataset \cite{5}; LSTMs on Penn Treebank dataset \cite{32}.

\subsection{Details}
\paragraph{Hyperparameters}
For SGDM and AEGD, we employ the grid search for learning rates in $\{1, 0.5, 0.3, 0.1, 0.01\}$. We set momentum $\gamma$ in SGDM to the default value $0.9$. Note that reported in \cite{23}, the best learning rate for SGDM is $30$ for LSTMs on Penn Treebank dataset, and we follow this setting. For Adam, AMSGrad, AdaBound and AdaBelief, we employ the grid search for learning rates in $\{0.1, 5e-2, 1e-2, 5e-3, 1e-3\}$. We turn over $\beta_1$ values of $\{0.9,0.99\}$ and $\beta_2$ values of $\{0.99, 0.999\}$. For other parameters in above Adam-type methods, we follow the setting reported in \cite{23,17} for achieving the best performance on CIFAR-10, CIFAR-100 and Penn Treebank dataset and use the default values for other experiments. For DecGD, we use the default value of hyperparameters, the default learning rate for CIFAR-10 and CIFAR-100 and a warm-up learning rate for LSTMs. 

\begin{figure*}[ht]
	\centering
	\subfigure[different $c$ in DecGD]{
		\label{1a}
		\begin{minipage}[t]{0.32\linewidth}
		\includegraphics[width=\linewidth]{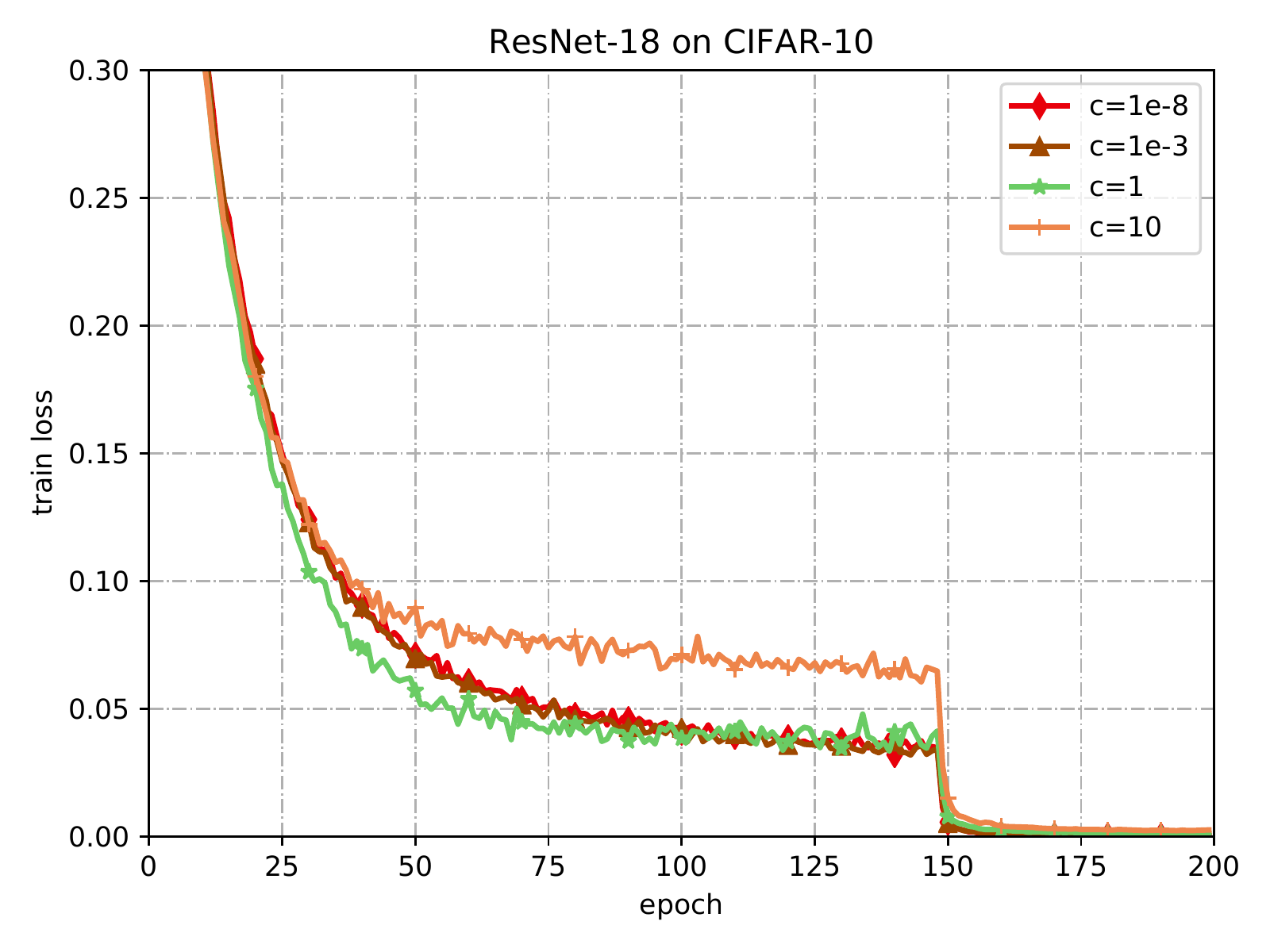}
	\end{minipage}}
	\subfigure[MLP on MNIST]{
		\label{2a}
		\begin{minipage}[t]{0.32\linewidth}
		\includegraphics[width=\linewidth]{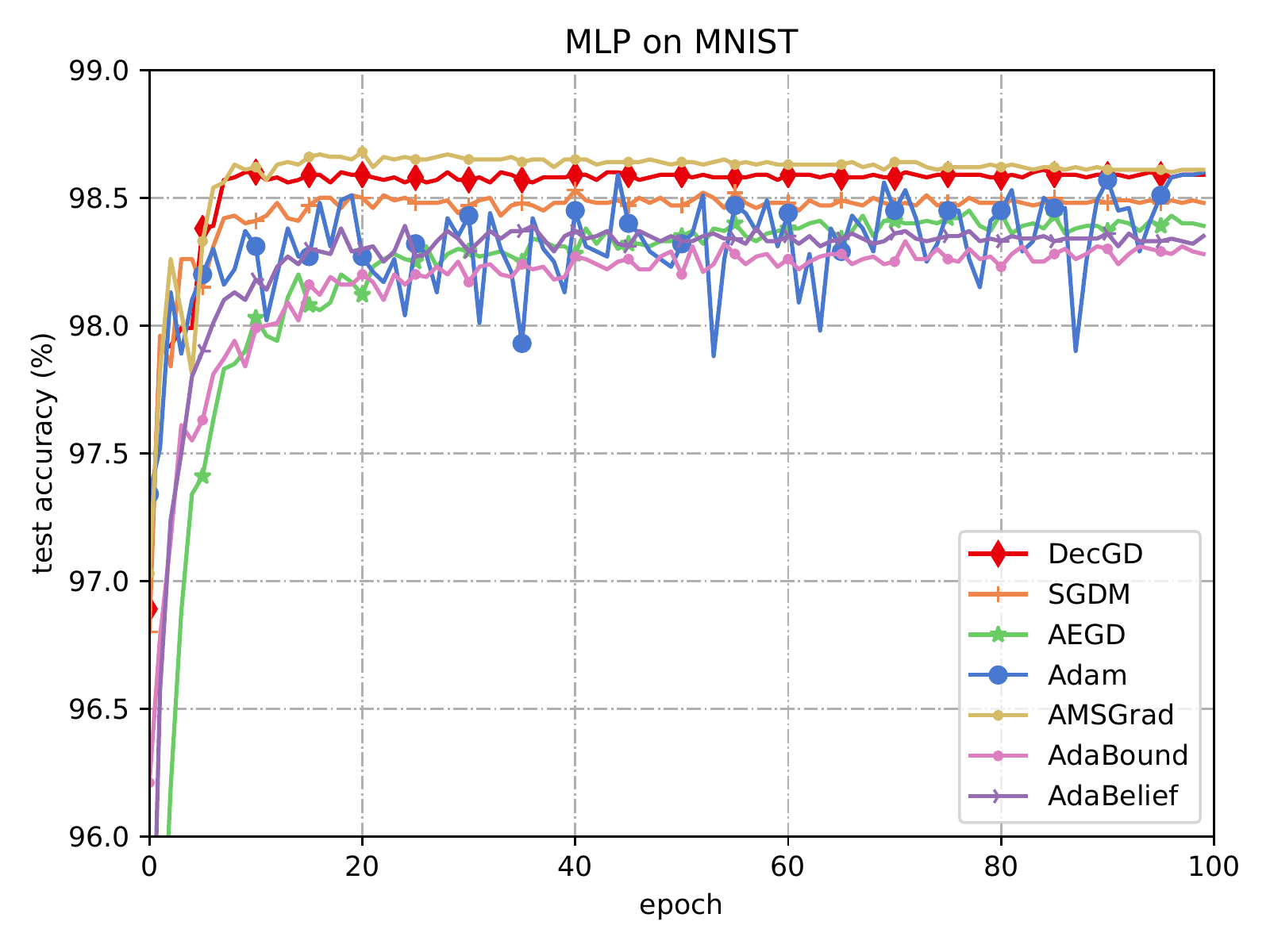}
	\end{minipage}}
	\subfigure[ResNet-34 on CIFAR-10]{
		\label{2b}
		\begin{minipage}[t]{0.32\linewidth}
		\includegraphics[width=\linewidth]{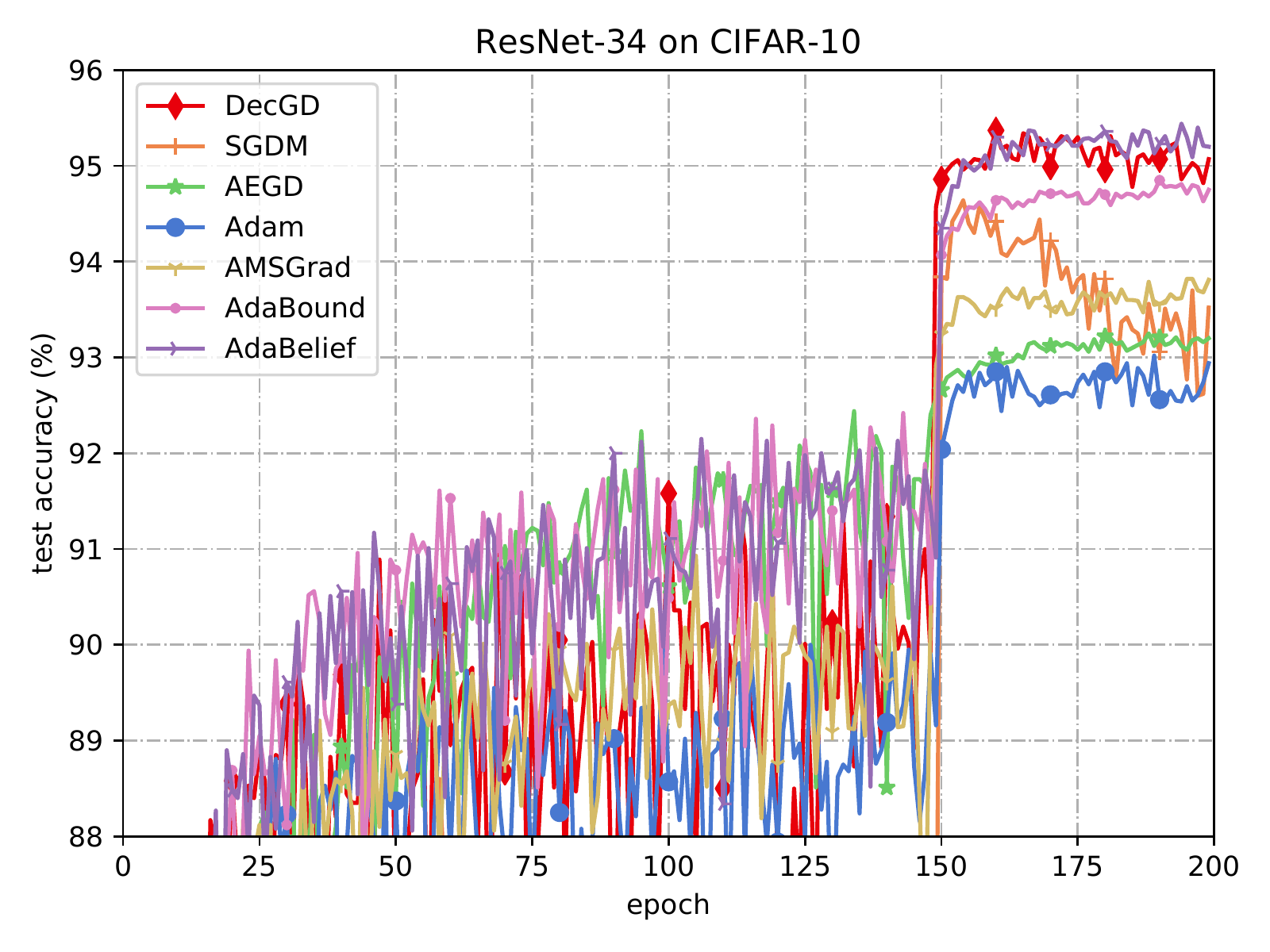}
	\end{minipage}}
	\subfigure[DesNet-121 on CIFAR-10]{
		\label{2c}
		\begin{minipage}[t]{0.32\linewidth}
		\includegraphics[width=\linewidth]{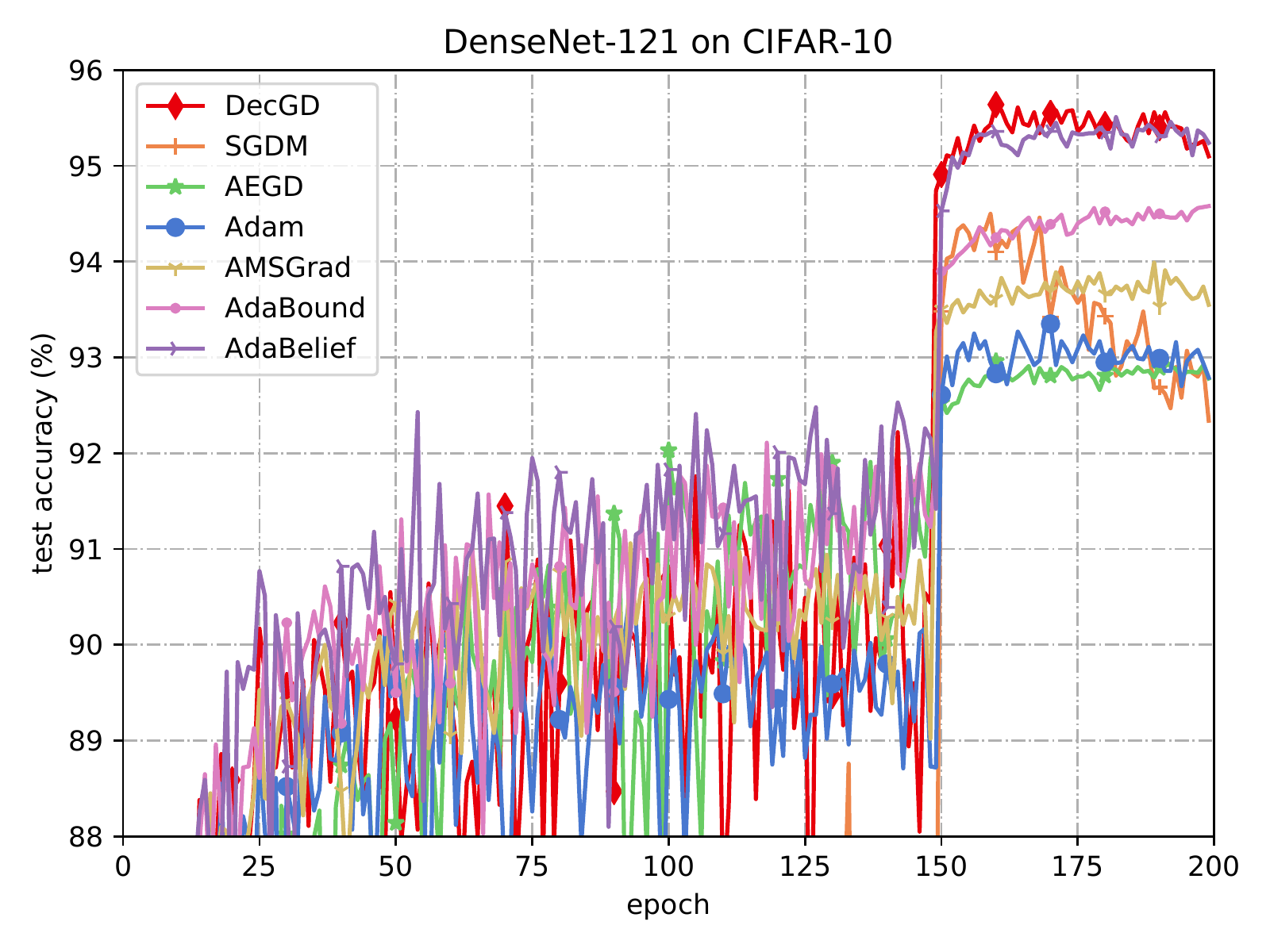}
	\end{minipage}}
	\subfigure[ResNet-34 on CIFAR-100]{
		\label{2d}
		\begin{minipage}[t]{0.32\linewidth}
		\includegraphics[width=\linewidth]{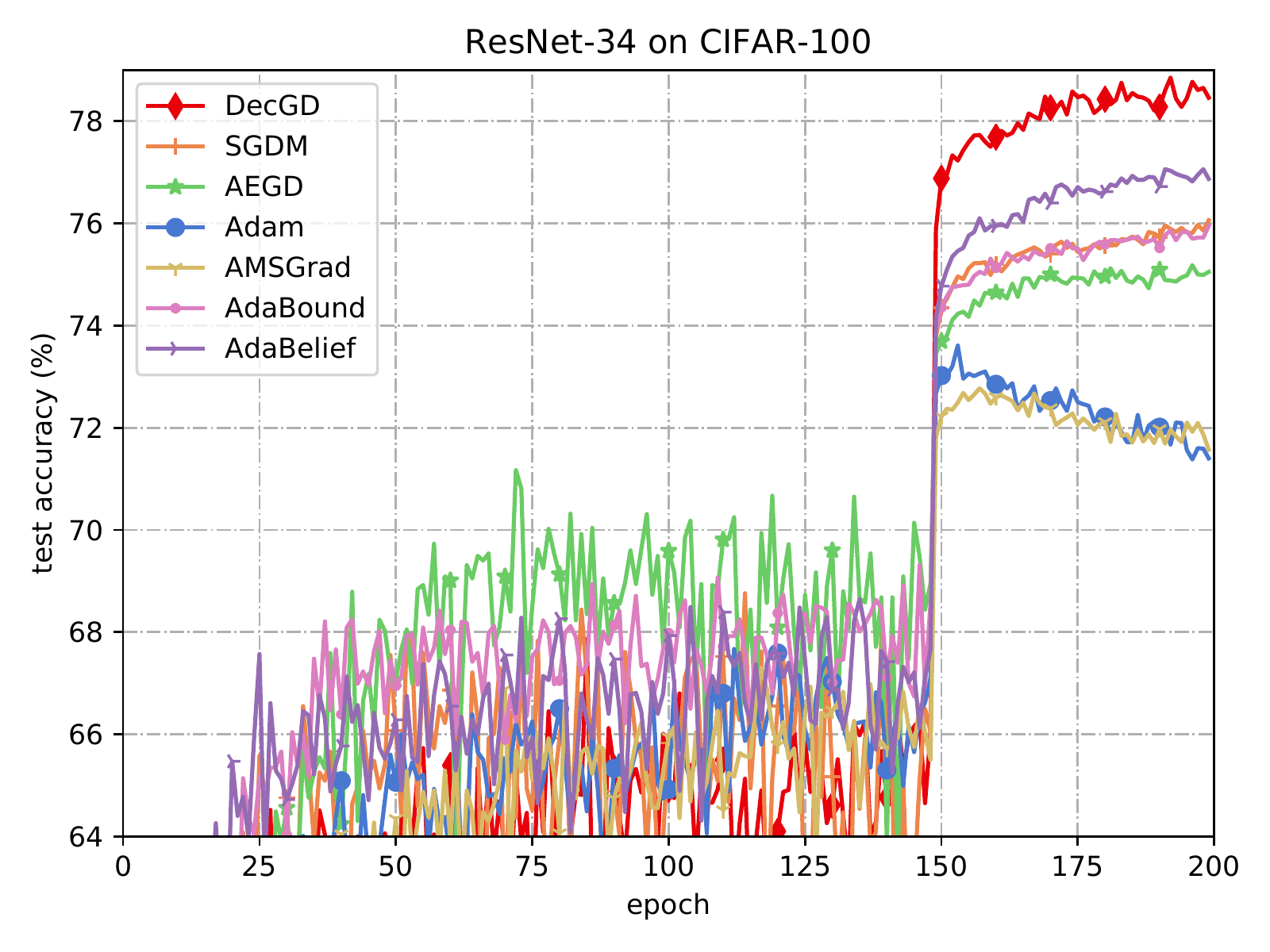}
	\end{minipage}}
	\subfigure[DesNet-121 on CIFAR-100]{
		\label{2e}
		\begin{minipage}[t]{0.32\linewidth}
		\includegraphics[width=\linewidth]{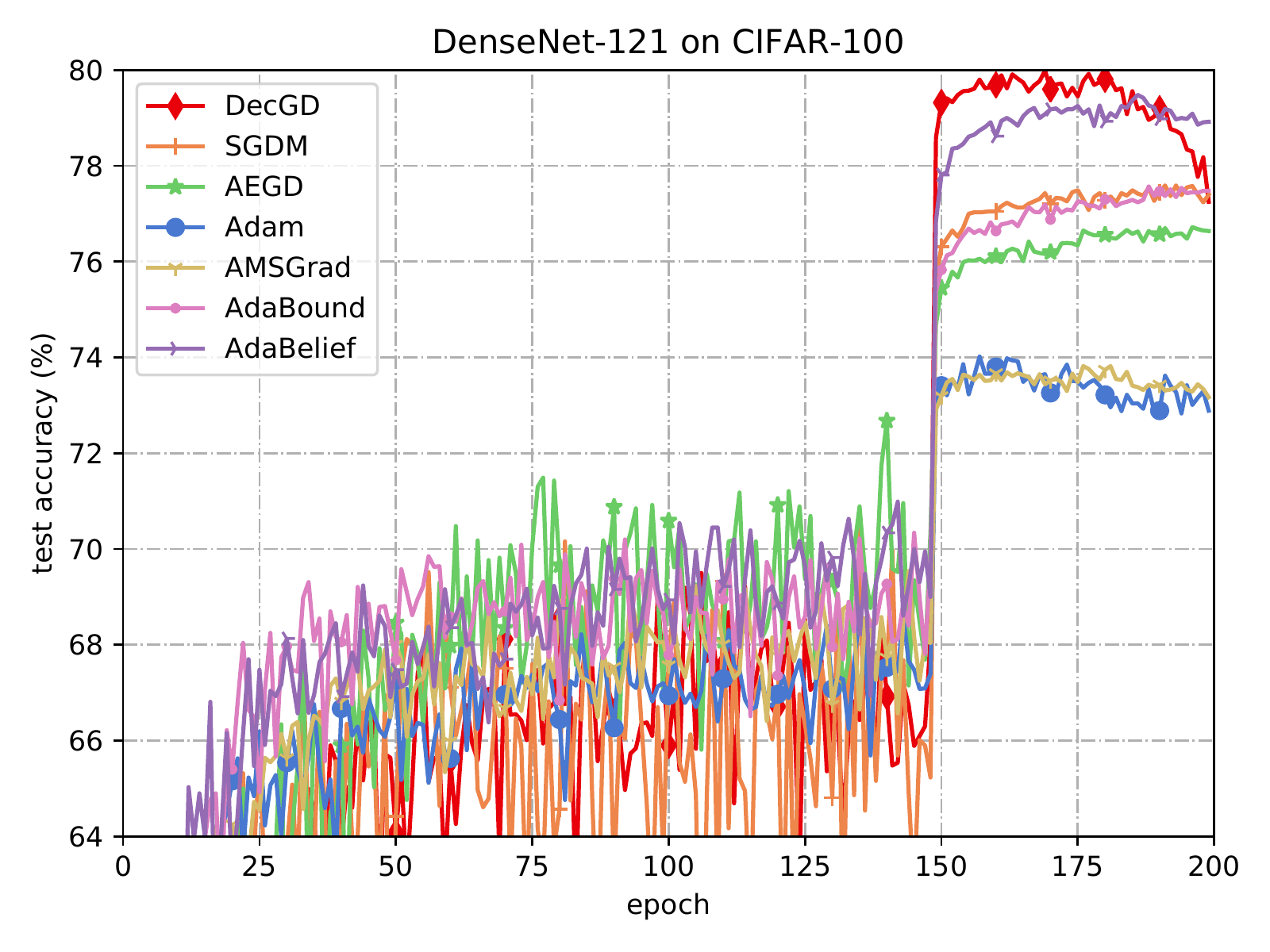}
	\end{minipage}}
	\subfigure[one layer LSTM on PTB]{
		\label{3a}
		\begin{minipage}[t]{0.32\linewidth}
		\includegraphics[width=\linewidth]{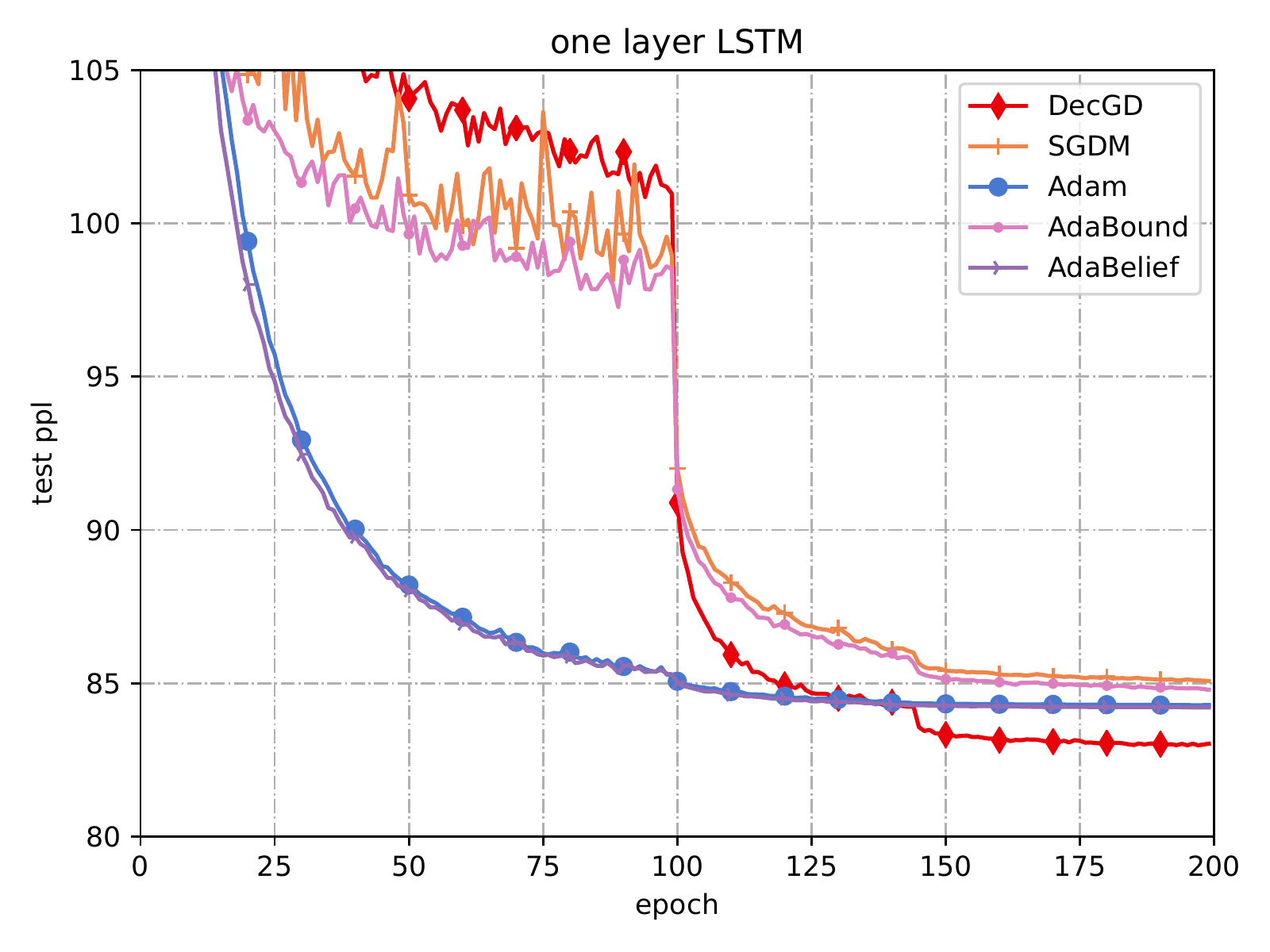}
	\end{minipage}}
	\subfigure[two layers LSTM on PTB]{
		\label{3b}
		\begin{minipage}[t]{0.32\linewidth}
		\includegraphics[width=\linewidth]{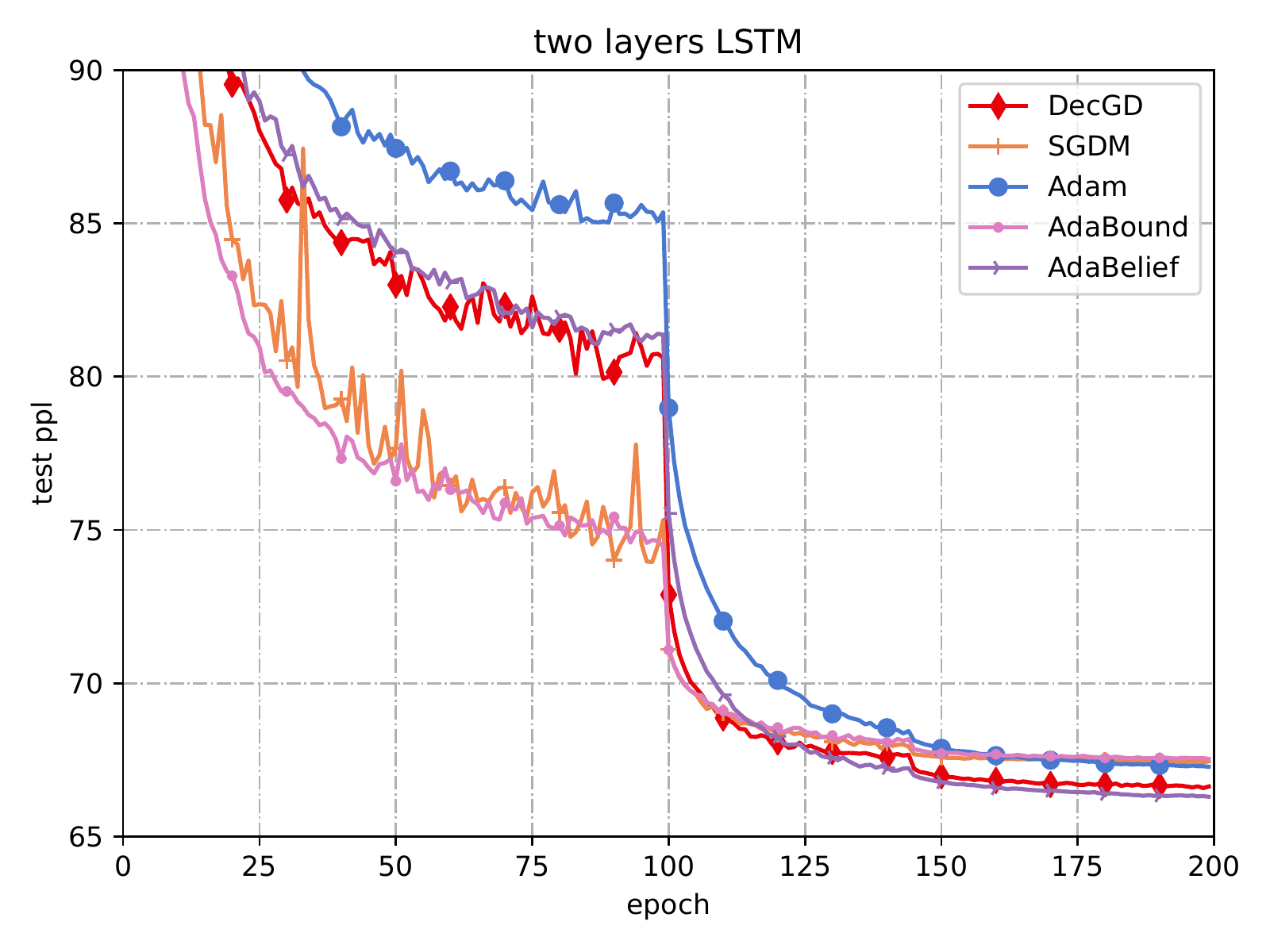}
	\end{minipage}}
	\subfigure[three layers LSTM on PTB]{
		\label{3c}
		\begin{minipage}[t]{0.32\linewidth}
		\includegraphics[width=\linewidth]{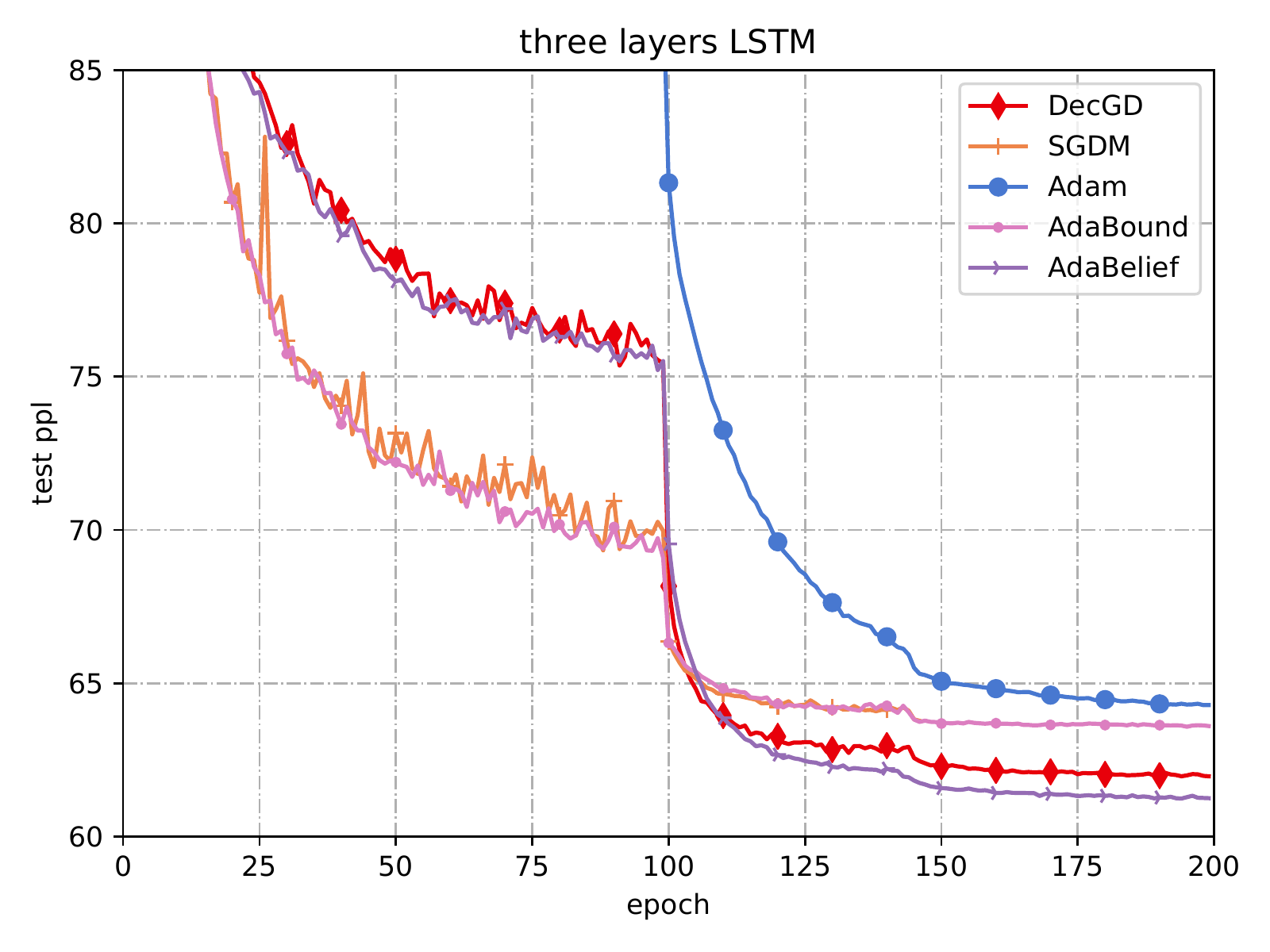}
	\end{minipage}}
	\label{fig 2}
	\caption{Performance of various optimizers in popular deep learning tasks. In (a)-(f), the higher is better and in (g)-(i), the lower is better. The results in (c) (d) (g) (h) (i), except AEGD, AMSGrad and our method DecGD, are reported in AdaBelief.}
\end{figure*}


  \begin{table}
	\caption{Test accuracy on CIFAR-10}
	\label{tab2}
	\centering
	\begin{tabular}{llllllll}
	  \toprule
	%
	  \cmidrule(r){1-2}
	  &SGDM     & AEGD     & Adam  &AMSGrad & AdaBound & AdaBelief & DecGD\\
	  \midrule
	  ResNet-34 & \makecell[c]{$94.64$}  & \makecell[c]{$93.25$} &\makecell[c]{$93.02$} & \makecell[c]{$93.82$} &\makecell[c]{$94.85$} &\makecell[c]{$\textbf{95.44}$}&\makecell[c]{$95.37$}\\
	  DenseNet-121&\makecell[c]{$94.5$} & \makecell[c]{$92.97$}  & \makecell[c]{$93.35$} &\makecell[c]{$93.99$} & \makecell[c]{$94.58$} &\makecell[c]{$95.51$} &\makecell[c]{$\textbf{95.64}$}\\
	  \bottomrule
	\end{tabular}
  \end{table}

  \begin{table}
	\caption{Test accuracy on CIFAR-100}
	\label{tab3}
	\centering
	\begin{tabular}{llllllll}
	  \toprule
	%
	  \cmidrule(r){1-2}
	  &SGDM     & AEGD     & Adam  &AMSGrad & AdaBound & AdaBelief & DecGD\\
	  \midrule
	  ResNet-34 & \makecell[c]{$76.06$}  & \makecell[c]{$75.18$} &\makecell[c]{$73.61$} & \makecell[c]{$72.77$} &\makecell[c]{$75.95$} &\makecell[c]{$77.06$}&\makecell[c]{$\textbf{78.85}$}\\
	  DenseNet-121&\makecell[c]{$77.59$} & \makecell[c]{$76.72$}  & \makecell[c]{$74.02$} &\makecell[c]{$73.82$} & \makecell[c]{$77.57$} &\makecell[c]{$79.48$} &\makecell[c]{$\textbf{79.99}$}\\
	  \bottomrule
	\end{tabular}
  \end{table}

  \begin{table}
	\caption{Test perplexity of LSTMs on PTB}
	\label{tab4}
	\centering
	\begin{tabular}{lllllll}
	  \toprule
	%
	  \cmidrule(r){1-2}
	  &SGDM     & Adam  & AdaBound & AdaBelief & DecGD\\
	  \midrule
	  1-layer & \makecell[c]{$85.07$}  &\makecell[c]{$84.28$} & \makecell[c]{$84.78$} &\makecell[c]{$84.20$} &\makecell[c]{$\textbf{82.97}$}\\
	  2-layer & \makecell[c]{$67.42$}  &\makecell[c]{$67.27$} & \makecell[c]{$67.52$} &\makecell[c]{$\textbf{66.29}$} &\makecell[c]{$66.57$}\\
	  3-layer & \makecell[c]{$63.77$}  &\makecell[c]{$64.28$} & \makecell[c]{$63.58$} &\makecell[c]{$\textbf{61.23}$} &\makecell[c]{$61.96$}\\
	  \bottomrule
	\end{tabular}
  \end{table}

\paragraph{MLP on MNIST}
We conduct the experiment to test the performance of aforementioned optimizers with MLP on MNIST. We follow the experiment settings reported in AdaBound \cite{17}. MLP is a fully connected neural network with only one hidden layer and total epoch is $100$. Figure \ref{2a} shows the empirical result. Note that all optimization algorithms achieve a test error below $2\%$ and our method DecGD and AMSGrad achieve slightly better performance than other methods on the test set.

\paragraph{ResNet-34 and DenseNet-121 on CIFAR-10}
CIFAR-10 is a more complex dataset than MNIST. We use more advanced and powerful deep convolution neural networks, including ResNet-34 and DenseNet-121, to test various optimization methods in this classification task on CIFAR-10 dataset. We employ the fixed budget of 200 epochs, set the mini-batch size to $128$. Figure \ref{2b} and \ref{2c} show the empirical results. Code is modified from the official implementation of AdaBelief and except AEGD and DecGD, data of other optimizers are reported in \cite{23}. As expected, the overall performances of each algorithm on ResNet-34 are similar to those on DenseNet-121. We note that DecGD shows the best generalization performance on DenseNet-121. For ResNet-34, the error of DecGD is slightly lower than that of AdaBelief with a margin $0.07\%$ which is the best performance; For DenseNet-121, DecGD surpasses AdaBelief with a margin $0.13\%$. We find that classical adaptive methods show rapid descent in the early period of training such as Adam and AdaBound. However, they show mediocre generalization ability on the test set. The empirical results show that our method overcomes the above drawback and achieves even better generalization performance than SGDM.

\paragraph{ResNet-34 and DenseNet-121 on CIFAR-100}
CIFAR-100 is similar to CIFAR-10, but the total class is up to $100$. Therefore, CIFAR-100 is more difficult and more close to the reality than CIFAR-10. We choose the same structures ResNet-34 and DenseNet-121. We employ the fixed budget of 200 epochs, set the mini-batch size to $128$. Figure \ref{2d} and \ref{2e} show the empirical results. Code is modified from the official implementation of AdaBelief. Note that DecGD achieves the best generalization performance both in ResNet-34 and DenseNet-121. Concretely, DecGD surpasses AdaBelief with a margin $1.79\%$ in ResNet-34 and a margin $0.51\%$ in DenseNet-121. DecGD shows the generalization ability far beyond other methods on CIFAR-100.

\paragraph{Robustness to $c$ change}
Considering the popularity of classification tasks, we test the performance of different $c$ for better application of DecGD in these tasks. We select ResNet-18 on CIFAR-10 dataset and $c$ is chosen from $\{10,1,1e-3,1e-8\}$. Figure \ref{1a} shows the result. Note that DecGD is robust to different $c$ and the default value $c=1$ achieves the slightly better performance. As a sequence, we use the default $c=1$ for almost all deep learning experiments.

\paragraph{LSTMs on Penn Treebank dataset}
We test our method on Penn Treebank dataset with one-layer LSTM, two-layers LSTM and three-layers LSTM respectively. We follow the setting of experiments in AdaBelief \cite{23}. One difference is that AdaBelief improves these experiments by setting learning rate scheduling at epoch $100$ and $145$ in their official implementation. Except our methods, the results of other methods are reported in AdaBelief. Code is modified from the official implementation of AdaBelief. The perplexities (ppl) of methods are reported in Figure \ref{3a}, \ref{3b} and \ref{3c} except AEGD and AMSGrad due to their worse performances. 
To our knowledge, AdaBelief has been the best optimizer on Penn Treebank dataset. Note that our method DecGD achieves the similar performance to AdaBelief in all three experiments: For one-layer LSTM, DecGD surpasses AdaBelief and achieves the lowest perplexity; For two-layers LSTM, DecGD and AdaBelief show the best performance and DecGD is higher than AdaBelief by a margin $0.28$; For three-layers LSTM, DecGD shows the lower perplexity than other methods except AdaBelief and is higher than AdaBelief by a margin $0.73$. Considering perplexities equal to $e^{loss}$, the gap between AdaBelief and DecGD is very small.

\subsection{Analysis}
We select the popular tasks of computer vision and natural language processing to investigate the generalization performance of our proposed methods. As shown above, DecGD, as a new adaptive method, shows an excellent generalization performance which is even better than SGDM. Besides, DecGD is robust to hyperparameters and achieves the best performance with the default learning rate in most cases, especially on CIFAR-100 dataset.

\section{Conclusion}
We have introduced DecGD, a simple and computationally efficient adaptive algorithm for non-convex stochastic optimization. This method is aimed towards large-scale optimization problems in the sense of large datasets and/or high-dimensional parameter spaces such as machine learning and deep neural networks. The practical intuition and excellent performance of DecGD show that our method is worth further research.

Despite excellent performance of our method, there still remains several directions to explore in the future:
\begin{itemize}
	\item [$\bullet$] First, we prove a $O(\log T/\sqrt{T})$ bound of our method DecGD in non-convex setting. However, empirical results show that the generalization performance of DecGD is better than many methods with a similar bound such as Adam. A tighter regret bound of DecGD needs to be explored in the future.
	\item [$\bullet$] Furthermore, as mentioned before, DecGD and Adam can be integrated with each other. This topic remains open.
	\item [$\bullet$] Finally, several works aim to find a new way to generate adaptive learning rates different from Adam-type methods. Thus, as this kind of works increase, how to measure the quality of adaptive learning rates is more and more important. However, there are few works on this topic.
\end{itemize}

\section{Broader Impact}
Optimization is at the core of machine learning and deep learning. To our best knowledge, there exists few
works on different adaptivity from Adam-type methods or EMA mechanism. DecGD shows better performance than those methods. DecGD can boost the application in all models which can numerically estimate parameter gradients.


\bibliographystyle{plainnat}
\bibliography{neurips_2021}

\clearpage
\appendix
\appendixpage
\addappheadtotoc
\section{Proof of Lemma 1}

	
\begin{proof}
	Because $l_t(x_t)=\sqrt{f_t(x_t)+c}$, $x_t\in\mathcal{X}$, then
	\begin{equation*}
		\nabla f_t(x_t)=2l_t(x_t)\nabla l_t(x_t).
	\end{equation*}
	If $\nabla f_t(x_t)$ is bounded, $l_t(x_t)\nabla l_t(x_t)$ is bounded. Therefore, at least one of $l_t(x_t)$ and $\nabla l_t(x_t)$ is bounded. 
		
	First, consider $l_t(x_t)$ is bounded and $\nabla l_t(x_t)$ is unbounded, the only possible case is that
	\begin{equation*}
		l_t(x_t)\to 0,\ \ ||\nabla l_t(x_t)||\to \infty.
	\end{equation*}
	Obviously, this case doesn't exist because if $l_t(x_t)\to 0$, $l_t$ must have gradients with a limit of $0$. 
		
	Second, consider $\nabla l_t(x_t)$ is bounded and $l_t(x_t)$ is unbounded, the only possible case is that
	\begin{equation*}
		l_t(x_t)\to \infty,\ \ ||\nabla l_t(x_t)||\to 0.
	\end{equation*}
	However, if $l_t(x_t)$ has gradients with a limit of 0, $l_t(x_t)$ must be finite. Thus, this case doesn't exist too.
	
	Finally, the only case is that $l_t(x_t)$ is bounded and $\nabla l_t(x_t)$ is bounded too. This proves that if $f_t$ has bounded gradients, then $l_t$ has bounded gradients too and is bounded in the feasible regions.
\end{proof}

\section{Proof of Theorem 1}
\begin{proof}
	We first replace the element-wise product with the dialog matrix and obtain
	\begin{equation*}
		x_{t+1}=x_t-2\eta V_{t}m_t,
	\end{equation*}
	where $V_{t}=\mathrm{diag}\{v^*\}$. If AMS is true, $V_t$ is monotonic decreasing. We aim to minimize the following regret:
	\begin{equation*}
		R(T)=\sum_{i=1}^Tf_t(x_t)-\min_{x\in\mathcal{X}}\sum_{i=1}^Tf_t(x).
	\end{equation*}
	Let $x^*\in\mathcal{X}$ be the optimal solution, the above regret is
	\begin{equation*}
		\begin{aligned}
			R(T)&=\sum_{i=1}^Tf_t(x_t)-\sum_{i=1}^Tf_t(x^*)\\
			&=\sum_{i=1}^T(f_t(x_t)-f_t(x^*))\\
			&\le \sum_{i=1}^T\big<\nabla f_t(x_t), x_t-x^*\big>.        
		\end{aligned}
	\end{equation*}
	Because
	\begin{equation*}
		\begin{aligned}
			||V_{t+1}^{-\frac{1}{2}}(x_{t+1}-x^*)||^2
			\le&||V_{t}^{-\frac{1}{2}}(x_t-2\eta_t V_{t}m_{t}-x^*)||^2\\
			=&||V_{t}^{-\frac{1}{2}}(x_t-x^*)||^2+4\eta_t^2||V_{t}^{\frac{1}{2}}m_{t}||^2-4\eta_t\big<m_{t},x_t-x^*\big>\\
			=&||V_{t}^{-\frac{1}{2}}(x_t-x^*)||^2+4\eta_t^2||V_{t}^{\frac{1}{2}}m_{t}||^2-4\eta_t\big<\gamma m_{t-1}+\nabla l_t(x_t), x_t-x^*\big>\\
			=&||V_{t}^{-\frac{1}{2}}(x_t-x^*)||^2+4\eta_t^2||V_{t}^{\frac{1}{2}}m_{t}||^2-4\eta_t\gamma\big<m_{t-1},x_t-x^*\big>-4\eta_t\big<\nabla l_t(x_t), x_t-x^*\big>,
		\end{aligned}
	\end{equation*}
	we have
	\begin{equation*}
		\begin{aligned}
			\big<\nabla l_t(x_t), x_t-x^*\big>
			\le&\frac{1}{4\eta_t}\big(||V_{t}^{-\frac{1}{2}}(x_t-x^*)||^2-||V_{t}^{-\frac{1}{2}}(x_{t+1}-x^*)||^2\big)+\eta_t||V_{t}^{\frac{1}{2}}m_{t}||^2-\gamma\big<m_{t-1},x_t-x^*\big>\\
			\le&\frac{1}{4\eta_t}(||V_{t}^{-\frac{1}{2}}(x_t-x^*)||^2-||V_{t}^{-\frac{1}{2}}(x_{t+1}-x^*)||^2)+\eta_t||V_{t}^{\frac{1}{2}}m_{t}||^2\\
			&+\frac{\gamma^2}{\eta_t}||x_t-x^*||^2+\eta_t||m_{t-1}||^2,
		\end{aligned}
	\end{equation*}
	where the last inequality follows Cauchy-Schwarz inequality and Young's inequality. 
	
	Thus, we obtain
	\begin{equation*}
		\begin{aligned}
			\big<\nabla f_t(x_t), x_t-x^*\big>\le&\frac{l_t(x_t)}{4\eta_t}(||V_{t}^{-\frac{1}{2}}(x_t-x^*)||^2-||V_{t}^{-\frac{1}{2}}(x_{t+1}-x^*)||^2)+\eta_t l_t(x_t)||V_{t}^{\frac{1}{2}}m_{t}||^2
			\\&+\frac{\gamma^2 l_t(x_t)}{\eta_t}||x_t-x^*||^2+\eta_t l_t(x_t)||m_{t-1}||^2,
		\end{aligned}
	\end{equation*} 
	and the regret is as follows:
	\begin{equation*}
		\begin{aligned}
			R(T)\le&\sum_{t=1}^T\big<\nabla f_t(x_t), x_t-x^*\big>
			\\ \le&\sum_{t=1}^T\frac{l_t(x_t)}{4\eta_t}(||V_{t}^{-\frac{1}{2}}(x_t-x^*)||^2-||V_{t}^{-\frac{1}{2}}(x_{t+1}-x^*)||^2)+\sum_{t=1}^T\eta_t l_t(x_t)||V_{t}^{\frac{1}{2}}m_{t}||^2
			\\ &+\sum_{t=1}^T\frac{\gamma^2 l_t(x_t)}{\eta_t}||x_t-x^*||^2+\sum_{t=1}^T\eta_t l_t(x_t)||m_{t-1}||^2.
		\end{aligned}
	\end{equation*}
	We divide the right formula into three parts:
	\begin{equation*}
		\begin{aligned}
		&P_1=\sum_{t=1}^T\frac{l_t(x_t)}{4\eta_t}(||V_{t}^{-\frac{1}{2}}(x_t-x^*)||^2-||V_{t}^{-\frac{1}{2}}(x_{t+1}-x^*)||^2),\\
		&P_2=\sum_{t=1}^T\frac{\gamma^2 l_t(x_t)}{\eta_t}||x_t-x^*||^2,\\
		&P_3=\sum_{t=1}^T\eta_t l_t(x_t)||V_{t}^{\frac{1}{2}}m_{t}||^2+\sum_{t=1}^T\eta_t l_t(x_t)||m_{t-1}||^2.
		\end{aligned}
	\end{equation*}
	Consider the part $1$ and apply Assumption 2:
	\begin{equation*}
		\begin{aligned}
		P_1=&\sum_{t=1}^T\frac{l_t(x_t)}{4\eta_t}(||V_{t}^{-\frac{1}{2}}(x_t-x^*)||^2-||V_{t}^{-\frac{1}{2}}(x_{t+1}-x^*)||^2)\\
			\le&\frac{L}{4\eta}\sum_{t=1}^T\sqrt{t}\Big(||V_{t}^{-\frac{1}{2}}(x_t-x^*)||^2-||V_{t}^{-\frac{1}{2}}(x_{t+1}-x^*)||^2\Big)\\
			\le&\frac{L}{4\eta}\Big(||V_1^{-\frac{1}{2}}(x_1-x^*)||^2+\sum_{t=2}^T(\sqrt{t}||V_t^{-\frac{1}{2}}(x_t-x^*)||^2-\sqrt{t-1}||V_{t-1}^{-\frac{1}{2}}(x_t-x^*)||^2)\Big)\\
			\le&\frac{L}{4\eta}\Big(\sum_{i=1}^dv_{1,i}^{-1}(x_{1,i}-x_{i}^*)^2+\sum_{t=2}^T\sum_{i=1}^d(x_{t,i}-x^*_i)^2(\sqrt{t}v_{t,i}^{-1}-\sqrt{t-1}v_{t-1,i}^{-1})\Big)\\
			\le&\frac{L D^2}{4\eta}\Big(\sum_{i=1}^dv_{1,i}^{-1}+\sum_{t=2}^T\sum_{i=1}^d(\sqrt{t}v_{t,i}^{-1}-\sqrt{t-1}v_{t-1,i}^{-1})\Big)\\
			=&\frac{LD^2\sqrt{T}}{4\eta}\sum_{i=1}^dv_{T,i}^{-1}.
		\end{aligned}
	\end{equation*}
	Then, the part 2 is as follows:
	\begin{equation*}
		\begin{aligned}
		P_2=&\sum_{t=1}^T\frac{\gamma^2 l_t(x_t)}{\eta_t}||x_t-x^*||^2\le LD^2\sum_{t=1}^T\frac{\gamma^2}{\eta_t}.
		\end{aligned}
	\end{equation*}
	Finally, we give the upper bound of the part 3 by applying Assumption 2:
	\begin{equation*}
		\begin{aligned}
		P_3=&\sum_{t=1}^T\eta_t l_t(x_t)||V_{t}^{\frac{1}{2}}m_{t}||^2+\sum_{t=1}^T\eta_tl_t(x_t)||m_{t-1}||^2\\
		\le&\eta L\sum_{t=1}^T\frac{1}{\sqrt{t}}(||V_t^{\frac{1}{2}}m_t||^2+||m_{t-1}||^2)\\
		\le& \frac{\eta LG^2}{1-\gamma}\sum_{t=1}^T\frac{1}{\sqrt{t}}(v_t+1)\\
		\le&\frac{\eta LG^2(1+v_0)}{1-\gamma}\sum_{t=1}^T\frac{1}{\sqrt{t}}\\
		\le&\frac{\eta G^2L(1+L)}{1-\gamma}(2\sqrt{T}-1)
		\end{aligned}
	\end{equation*}

	Hence, we get the final regret bound:
	\begin{equation*}
		\begin{aligned}
			R(T)\le & \frac{LD^2\sqrt{T}}{4\eta}\sum_{i=1}^dv_{T,i}^{-1}+LD^2\sum_{t=1}^T\frac{\gamma^2}{\eta_t}+\frac{\eta G^2L(1+L)}{1-\gamma}(2\sqrt{T}-1).
		\end{aligned}
	\end{equation*}
\end{proof}

\section{Proof of Corollary 1}
\begin{proof}
	The inequalities $\sum_{t=1}^T\lambda^{2(t-1)}\sqrt{t}\le\sum_{t=1}^T\lambda^{2(t-1)}t\le\frac{1}{(1-\lambda^2)^2}$ yield
	\begin{equation*}
		\begin{aligned}
			R(T)\le\frac{LD^2\sqrt{T}}{4\eta}\sum_{i=1}^dv_{T,i}^{-1}+\frac{LD^2\gamma^2}{\eta(1-\lambda^2)^2}+\frac{\eta G^2L(1+L)}{1-\gamma}(2\sqrt{T}-1).
		\end{aligned}
	\end{equation*}	
\end{proof}
\section{Proof of Theorem 2}
We follow the proof in [3] and recall the Theorem 3.1 in [3]
\paragraph{Theorem 3.1}
[3] For an Adam-type method under the following assumptions:
\begin{itemize}
	\item [$\bullet$] $f$ is lower bounded and differentiable; $||\nabla f(x)-\nabla f(y)||\le L||x-y||,\forall x,y$.
	\item [$\bullet$] Both the true and noisy gradients are bounded, i.e. $||\nabla f(\theta)||\le H,||g_t||\le H,\forall t$.
	\item [$\bullet$] Unbiased and independent noise in $g_t$, i.e. $g_t=\nabla f(\theta_t)+\zeta_j$, $\mathbb{E}[\zeta_j]=0$, and $\zeta_i\perp\zeta_j$, $\forall i\neq j$.
\end{itemize}
Assume $\beta_t\le\beta\le1$ in non-increasing, $||\eta_tm_t/\sqrt{v_t}||\le G$, then:
\begin{equation*}
	\begin{aligned}
		&\mathbb{E}\Big(\sum_{t=1}^T\eta_t\big<\nabla f(\theta_t), \nabla f(\theta_t)/D_t\big>\Big)\\&\le \mathbb{E}\Big(C_1\sum_{t=1}^T\Big|\Big|\frac{\eta_tg_t}{\sqrt{v_{t}}}\Big|\Big|^2+C_2\sum_{t=2}^{T}\Big|\Big|\frac{\eta_t}{\sqrt{v_t}}-\frac{\eta_{t-1}}{\sqrt{v_{t-1}}}\Big|\Big|_1+C_3\sum_{t=2}^{T-1}\Big|\Big|\frac{\eta_t}{\sqrt{v_t}}-\frac{\eta_{t-1}}{\sqrt{v_{t-1}}}\Big|\Big|^2\Big)+C_4, 		
	\end{aligned}
\end{equation*}
where $C_1$, $C_2$, $C_3$ are constants independent of $d$ and $T$, $C_4$ is a constant independent of $T$.

We note that Theorem 3.1 in [3] gives the convergence bound for generalize Adam [3]. However, this general framework needs no EMA mechanism, i.e. squared gradients and represents more general adaptivity which uses first order gradients. DecGD belongs to this general framework with $1/\sqrt{v_t}$ in this general framework corresponding to $v_t$ in DecGD. Therefore, we can apply the above theorem to our proof of DecGD.
\begin{proof}
	According to the above theorem and Assumption 4, we obtain
	\begin{equation*}
		\begin{aligned}
			&\mathbb{E}\Big(\sum_{t=1}^T\eta_t\big<\nabla l(\theta_t),V_t \nabla l(\theta_t)\big>\Big)\\&\le \mathbb{E}\Big(C_1\sum_{t=1}^T\Big|\Big|\eta_tV_t\hat{g}_t\Big|\Big|^2+C_2\sum_{t=2}^{T}\Big|\Big|v_t\eta_t-v_{t-1}\eta_{t-1}\Big|\Big|_1+C_3\sum_{t=2}^{T-1}\Big|\Big|v_t\eta_t-v_{t-1}\eta_{t-1}\Big|\Big|^2\Big)+C_4, 		
		\end{aligned}
	\end{equation*}
	where $C_1$, $C_2$, $C_3$ are constants independent of $d$ and $T$, $C_4$ is a constant independent of $T$.

	We divide the right formula to three parts:
	\begin{equation*}
		\begin{aligned}
		&P_1=\mathbb{E}\sum_{t=1}^T\Big|\Big|\eta_tV_t\hat{g}_t\Big|\Big|^2,\\
		&P_2=\mathbb{E}\sum_{t=2}^{T}\Big|\Big|v_t\eta_t-v_{t-1}\eta_{t-1}\Big|\Big|_1,\\
		&P_3=\mathbb{E}\sum_{t=2}^{T-1}\Big|\Big|v_t\eta_t-v_{t-1}\eta_{t-1}\Big|\Big|^2.
		\end{aligned}
	\end{equation*}

	First, we give the upper bound of the part 1 according to Assumption 4:
	\begin{equation*}
		\begin{aligned}
			P_1\le L^2\eta^2\mathbb{E}\sum_{t=1}^T\Big|\Big|\frac{\hat{g}_t}{\sqrt{t}}\Big|\Big|^2\le L^2\eta^2G^2(1+\log T),
		\end{aligned}
	\end{equation*}
	where the last inequality is due to $\sum_{t=1}^T1/t\le1+\log T$.
	
	Then, consider the part 2:
	\begin{equation*}
		\begin{aligned}
			P_2=&\mathbb{E}\sum_{t=2}^{T}v_{t-1}\eta_{t-1}-v_t\eta_t\\
			=&\mathbb{E}\sum_{j=1}^d\eta_1 v_{1,j} -\eta_Tv_{T,j}\\
			\le&\mathbb{E}\sum_{j=1}^d\eta_1 v_{1,j}\\
			\le&dL\eta.
		\end{aligned}
	\end{equation*}

	Finally, the part 3 is as follows:
	\begin{equation*}
		\begin{aligned}
			P_3=&\mathbb{E}\sum_{t=2}^{T-1}||\eta_t v_t-\eta_{t-1}v_{t-1}||_1||\eta_t v_t-\eta_{t-1}v_{t-1}||_1\\
			\le& L\eta\mathbb{E}\sum_{t=2}^{T-1}||\eta_t v_t-\eta_{t-1}v_{t-1}||_1\\
			\le& dL^2\eta^2.
		\end{aligned}
	\end{equation*}

	Hence, we obtain
	\begin{equation*}
		\begin{aligned}
			&\mathbb{E}\Big(C_1\sum_{t=1}^T\Big|\Big|\eta_tV_t\hat{g}_t\Big|\Big|^2+C_2\sum_{t=2}^{T}\Big|\Big|v_t\eta_t-v_{t-1}\eta_{t-1}\Big|\Big|_1+C_3\sum_{t=2}^{T-1}\Big|\Big|v_t\eta_t-v_{t-1}\eta_{t-1}\Big|\Big|^2\Big)+C_4\\
			&\le C_1L^2\eta^2G^2(1+\log T)+C_2dL\eta+C_3 dL^2\eta^2+C_4		
		\end{aligned}
	\end{equation*}

	Now we lower bound the LHS. With the assumption $||v_T||_{1}\ge c$, we have
	\begin{equation*}
		(\eta_tV_t)_j=\frac{\eta (V_t)_j}{\sqrt{t}}\ge\frac{\eta c}{\sqrt{t}},
	\end{equation*}
	and thus
	\begin{equation*}
		\begin{aligned}
			\mathbb{E}\Big(\sum_{t=1}^T\eta_t\big<\nabla l(\theta_t),V_t \nabla l(\theta_t)\big>\Big)&\ge \mathbb{E}\Big(\sum_{t=1}^T\frac{\eta c}{\sqrt{t}}||\nabla l(\theta_t)||^2\Big)\ge\frac{\eta c}{L^2}\mathbb{E}\sum_{t=1}^T||\nabla f(\theta_t)||^2\\
			&\ge \frac{\eta c\sqrt{T}}{L^2}\min_{t\in[T]}\mathbb{E}||\nabla f(\theta_t)||^2.
		\end{aligned}
	\end{equation*}

	We finally obtain:
	\begin{equation*}
		\begin{aligned}
			\min_{t\in[T]}\mathbb{E}\Big(||\nabla f(\theta_t)||^2\Big)\le&\frac{L^2}{c\eta\sqrt{T}}\Big(C_1L^2\eta^2G^2(1+\log T)+C_2dL\eta+C_3dL^2\eta^2+C_4\Big)\\
			\le &\frac{1}{\sqrt{T}}(Q_1+Q_2\log T)
		\end{aligned}
	\end{equation*}
	This proves Theorem 2.
\end{proof}

\section{DecGD with Decoupled Weight Decay}

\begin{algorithm}
	\caption{DecGD with decoupled weight decay (default initialization: $c=1$, $\gamma=0.9$, $\eta=0.01$, AMS=False, $\lambda=1e-4$)}
	\label{alog2}
	\begin{algorithmic}
		\STATE{\bfseries Input:} $x_1 \in \mathcal{X}\subseteq\mathbb{R}^d$, learning rate $\eta$, $c$, momentum $\gamma$, AMS, decoupled weight decay $\lambda$
		\STATE{\bfseries Initialize} $x_{0,i}=0$, $m_{0,i}=0$, $v_{0,i}=\sqrt{f(x_1)+c}$, $i=1,2,\cdots,d$
		\FOR{$t=1$ {\bfseries to} $T$}
		\STATE $\nabla g\leftarrow \nabla f(x_t)/2\sqrt{f(x_t)+c}$
		\STATE $\nabla g\leftarrow \nabla g + \lambda x_t$
		\STATE $m_{t}\leftarrow \gamma m_{t-1} + \nabla g$
		\STATE $v_{t}\leftarrow v_{t-1} + m_t(x_{t}-x_{t-1})$
		\STATE $v^*=\min\{v^*,v_t\}$ \textbf{if AMS else} $v^*=v_t$
		\STATE $x_{t+1}\leftarrow x_{t}-\eta (2v^*m_t + \lambda x_t)$
		\ENDFOR
	\end{algorithmic}
\end{algorithm}

\section{Details of Experiments}
In this section, we give the details of the experiments reported in the body and show the extensive experiments for better understanding of our method DecGD.

\subsection{Details of Classification Tasks}
\paragraph{MNIST}
We employ a three-layers full connected neural network to test the performance of various methods. The batch size is $128$ and there is no trick applied to various methods due to the low difficulty of this task. The total epoch is 100.
\paragraph{CIFAR-10}
We follow the setting reported in AdaBound and AdaBelief. Moreover, the results of methods except AEGD and DecGD is exactly equal to the results reported in AdaBelief. For AEGD, we search the best learning rate in $[0.1,0.05,0.01,0.005,0.001]$ and use the recommended hyperparameters. For our method DecGD, all hyperparameters including the learning rate is default without tuning. The weight decay factor is $5e-4$ which is consistent with AdaBound and AdaBelief. We employ the milestone learning rate decay scheduling at epoch $150$, which is reported in AdaBelief and AdaBound. The total epoch is 200. This task can be viewed as the replication of experiments reported in AdaBelief, and we aim to achieve the fairest comparative experiments for all methods.
\paragraph{CIAFR-100}
The setting of two tasks is exactly same. We follow the setting reported in AdaBound and AdaBelief. Moreover, the results of methods except AEGD and DecGD is exactly equal to the results reported in AdaBelief. For AEGD, we search the best learning rate in $[0.1,0.05,0.01,0.005,0.001]$ and use the recommended hyperparameters. For our method DecGD, all hyperparameters including the learning rate is default without tuning. The weight decay factor is $5e-4$ which is consistent with AdaBound and AdaBelief. We employ the milestone learning rate decay scheduling at epoch $150$, which is reported in AdaBelief and AdaBound. The total epoch is 200. This task can be viewed as the replication of experiments reported in AdaBelief, and we aim to achieve the fairest comparative experiments for all methods.
\subsection{Details of Language Tasks}
\paragraph{Penn Treebank}
The experiments on Penn Treebank dataset are to test the performance of various methods with RNN. The models are one layer, two layers and three layers LSTM which are famous representatives of RNN. The setting is exactly same as that reported in AdaBelief. The total epoch is 200 and we employ the milestone learning rate decay scheduling at epoch $100$ and $145$ which is reported in the official implementation of AdaBelief. However, AMSGrad and AEGD show bad performance and need more efforts on tuning hyperparameters and learning rates. We will add the results of them in the future. For DecGD, considering the learning rate of SGD is up to $30$, and we follow this setting to use a big learning rate in early training and increase the default learning rate to $0.3$ for LSTMs. The weight decay factor we used is reported in AdaBelief.

\subsection{Robustness to Hyperparameters}
We test various $c$ in the body and report the setting of experiments. Robustness of other hyperparameters such as AMS and learning rates will be test in the future.

\paragraph{c}
Note that $c$ is the most important hyperparameter in DecGD. We test different $c$ in a popular task: ResNet-18 in CIFAR-10. The learning rate is default, and we follow the setting reported in AdaBelief including weight decay and learning rate scheduling. At least for zero-lower-bounded loss functions, DecGD is robust to $c$ change and $1$ is a good default initialization. We will test more functions with different lower bounds in the future.

\subsection{Numerical Experiments}
We conduct numerical experiments on the two classical high-dimensional functions, Extended Powell Singular function and Extended Rosenbrock function, to test the performance of our method in a convex and non-convex function respectively. The former is a high-dimensional convex function, but the Hessian has a double singularity at the solution so that in the global optimization literature this function is stated as a difficult test case. Extended Powell Singular function is as follows:
\begin{equation*}
	\begin{aligned}
		f(\mathbf{x})= & \sum_{i=1}^{N/4}\Big((x_{4i-3}+10x_{4i-2})^2+5(x_{4i-1}-x_{4i})^2+(x_{4i-2}-2x_{4i-1})^4+10(x_{4i-3}-x_{4i})^4\Big),
	\end{aligned}
\end{equation*}
where we set $N=100$ and the initial point is $(3,-1,0,1,3,-1,0,1,\cdots,3,-1,0,1)^T$. The exact optimal solution is $(0,0,\cdots,0)^T$ and the exact minimum value of Extended Powell Singular function is $0$. The latter, Extended Rosenbrock function, is a famous high-dimensional non-convex function in optimization. It has the following form:
\begin{equation*}
	f(\mathbf{x})=\sum_{i=1}^{N/2}\Big(100(x_{2i-1}^2-x_{2i})^2+(x_{2i}-1)^2\Big)
\end{equation*}
where we set $N=1000$. We start with initial point $x_0=(-1.2,1,-1.2,1,\cdots,-1.2,1)^T$ to find the optimal solution. The exact optimal solution is $(1,1,\cdots,1)^T$ and the exact minimum value of Extended Rosenbrock function is $0$. 

\begin{figure*}[t] 
	\centering  
	\subfigure[Extended Powell Singular function]{
	\label{fig f1}
	\begin{minipage}[t]{0.5\linewidth}
	\centering
	\includegraphics[width=\linewidth]{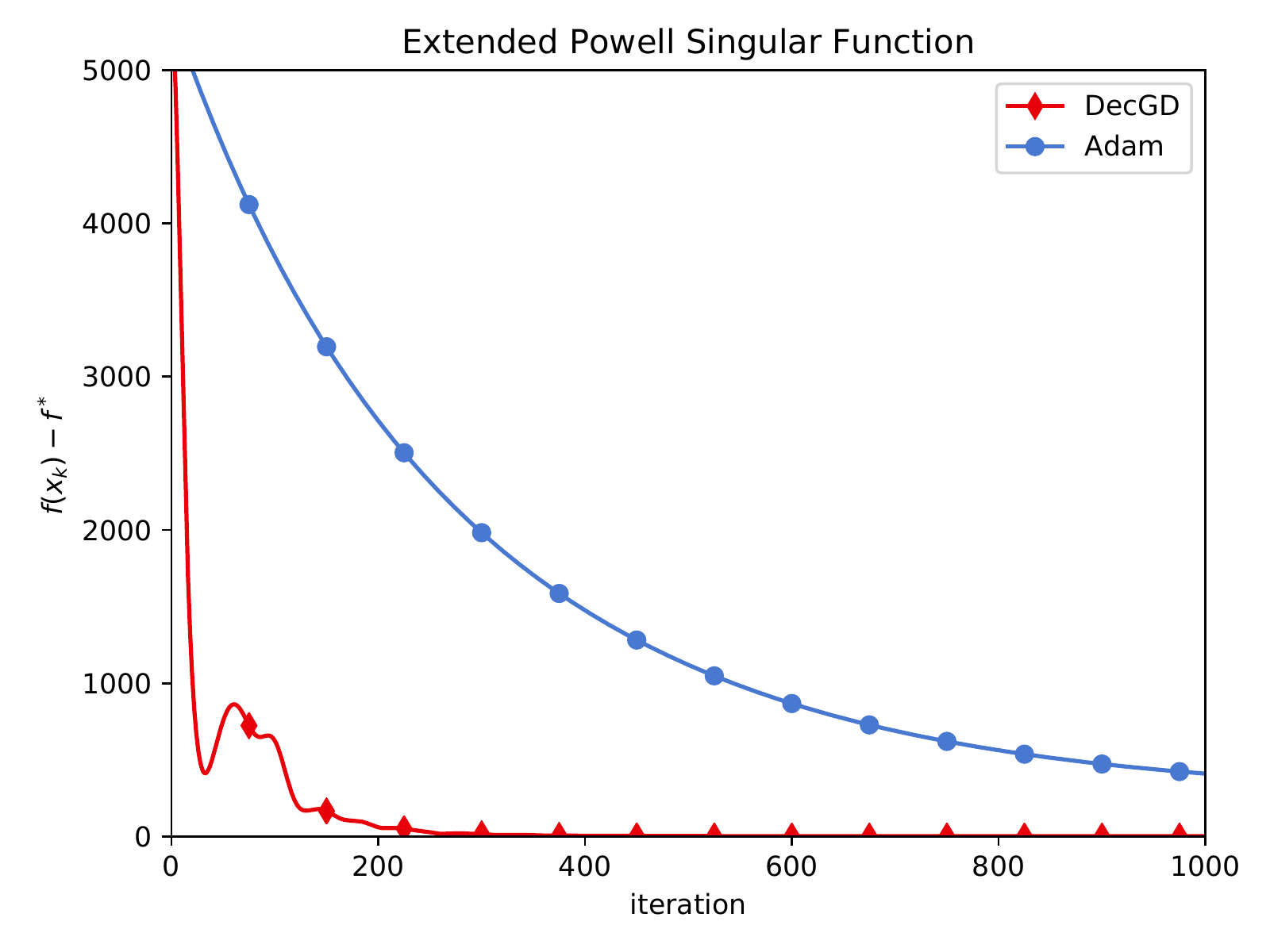}
	\end{minipage}%
	}%
	\subfigure[Extended Rosenbrock function]{
		\label{fig1(b)}
		\begin{minipage}[t]{0.5\linewidth}
		\centering
		\includegraphics[width=\linewidth]{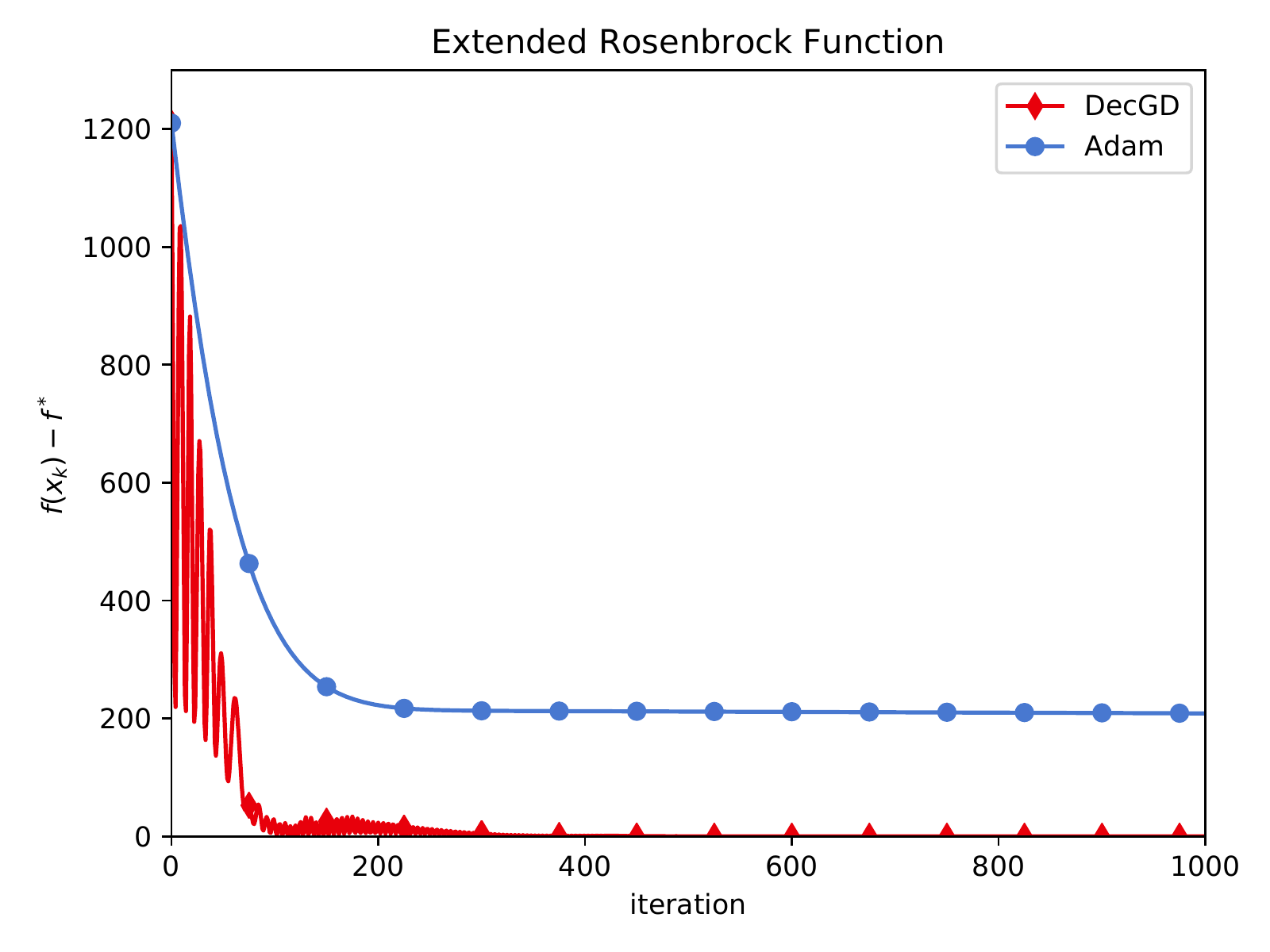}
		\end{minipage}%
		}%
	\caption{Performance comparison on Extended Powell Singular function and Extended Rosenbrock function.} 
	\label{fig1}
\end{figure*}

We test Adam and our method DecGD in these two functions and the learning rates are $1e-3$ for Adam and $1e-5$ for DecGD. The results are reported in Figure \ref{fig f1}. We note that DecGD converge faster than Adam with even smaller learning rates. However, DecGD shows more oscillation and zigzag than Adam. How to reduce them remains open.

\end{document}